\documentclass[10pt, twocolumn, letterpaper]{article}

\usepackage[pagenumbers]{cvpr} %

\usepackage{booktabs}
\usepackage{graphicx}
\usepackage{multirow}
\usepackage[table]{xcolor}
\usepackage{xspace}
\usepackage{enumitem}
\usepackage{subcaption}
\usepackage{dashrule}
\usepackage{algorithm}
\usepackage{algorithmicx}
\usepackage{algpseudocode}
\usepackage{url}
\usepackage{pifont}
\usepackage{float}

\definecolor{gain green}{RGB}{0,140,70}
\definecolor{gray}{RGB}{100,100,100}
\newcommand{\gain}[1]{\textcolor{gain green}{(#1)}}
\newcommand{\gray}[1]{\textcolor{gray}{#1}}

\newcommand{\method}{TILDE\xspace}
\newcommand{\ad}{AutoDAgger\xspace}

\newcommand{\cmark}{\ding{51}}
\newcommand{\xmark}{\ding{55}}

\usepackage{amsmath,amsfonts,bm}

\def\eqref#1{equation~\ref{#1}}

\def\1{\bm{1}}

\def\va{{\bm{a}}}

\def\vl{{\bm{l}}}

\def\vo{{\bm{o}}}
\def\vp{{\bm{p}}}
\def\vq{{\bm{q}}}

\def\vs{{\bm{s}}}

\def\mA{{\bm{A}}}

\DeclareMathAlphabet{\mathsfit}{\encodingdefault}{\sfdefault}{m}{sl}
\SetMathAlphabet{\mathsfit}{bold}{\encodingdefault}{\sfdefault}{bx}{n}

\definecolor{cvprblue}{rgb}{0.21,0.49,0.74}
\usepackage[pagebackref, breaklinks, colorlinks, allcolors=cvprblue]{hyperref}

\def\confName{CVPR}
\def\confYear{2027}

\title{TILDE: Teacher-in-the-Loop Distillation from Embodied Foundation Models}

\author{Yuxiang Lu\thanks{Equal Contribution} \enspace  Yantong Liu$^*$ \enspace  Jinghua Hou \enspace  Xueli Sun \enspace  Ruihua Han \enspace  Zhe Liu \enspace  Hengshuang Zhao\thanks{Corresponding Author}\\
The University of Hong Kong \\
{\tt\small https://hku-sail.github.io/TILDE}
}

\makeatletter
\g@addto@macro\@maketitle{ \vspace{-0.5cm} \begin{figure}[H]\setlength{\linewidth}{\textwidth} \setlength{\hsize}{\textwidth} \centering
\includegraphics[width=\linewidth]{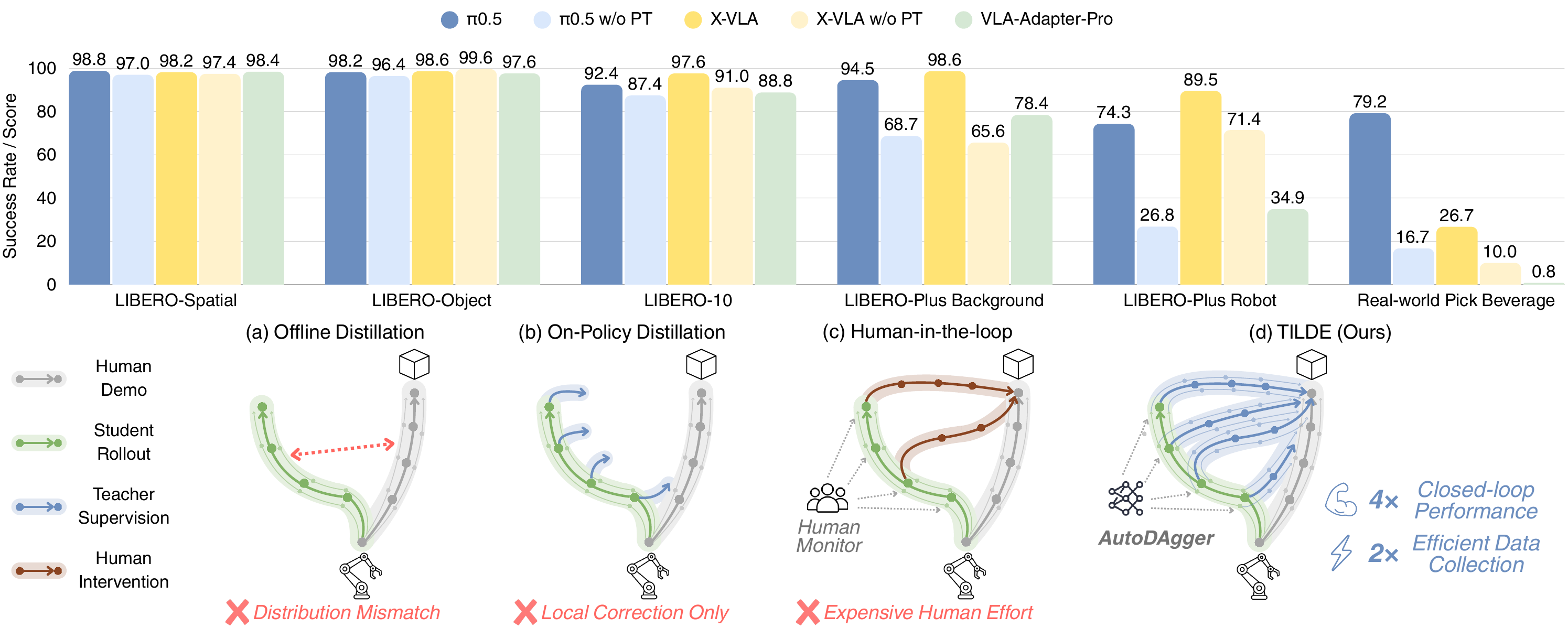} \caption{ \textbf{Top:} Policies with ($\pi_{0.5}$ \& X-VLA) and without large-scale action pretraining (``w/o PT'' \& VLA-Adapter-Pro) perform similarly on standard benchmarks (LIBERO-Spatial \& Object), while larger gaps emerge on long-horizon tasks (LIBERO-10), under perturbations (LIBERO-Plus), and on the real-world task. These results motivate transferring closed-loop capabilities from EFMs to student policies without repeating action pretraining. \textbf{Bottom:} (a) Offline distillation is restricted to the demonstration distribution, leading to distribution mismatch during student execution. (b) On-policy distillation extends supervision to student-induced states but does not execute teacher corrections to observe the subsequent states. (c) Human-in-the-loop methods collect recovery trajectories, but their coverage is limited by the bandwidth of human effort. (d) \method jointly learns from demonstration, student rollout, and teacher recovery experience to transfer closed-loop capability, with \ad automating progress monitoring and teacher intervention for efficient data collection. } \label{fig:teaser}\end{figure} }

\begin{document}
    \maketitle

    \begin{abstract}
    Large-scale action pretraining equips Embodied Foundation Models (EFMs)
    with strong manipulation capabilities, including generalization and closed-loop robustness, but
    requires substantial robot data and computation. We introduce TILDE
    (\textbf{T}eacher-\textbf{I}n-the-\textbf{L}oop \textbf{D}istillation from \textbf{E}FMs), a framework that transfers
    these capabilities to student policies without robot action pretraining.
    TILDE combines teacher supervision at student-visited states with
    learning from teacher-executed recovery trajectories, complementing
    offline human demonstrations. This exposes the student to both corrective
    action targets and the subsequent states reached during teacher
    recovery. To collect this experience, we propose AutoDAgger, an autonomous
    data aggregation pipeline that uses a robotic reward model to monitor
    task progress and an execution manager to trigger teacher takeover.
    By automating monitoring and intervention, AutoDAgger improves collection throughput while reducing
    reliance on human oversight and teleoperation.
    Experiments on real-world manipulation tasks and LIBERO benchmark demonstrate
    significantly improved student performance and effective cross-architecture transfer.
    \end{abstract}

    \section{Introduction}

    Embodied Foundation Models (EFMs), including Vision-Language-Action (VLA) models~\citep{zhang2025pure, shao2025large, ma2024survey, sapkota2025vision},
    have demonstrated remarkable manipulation and generalization capabilities through action pretraining on large-scale robot
    datasets~\citep{agibot_world, oxe, bridgedata, droid}.
    Reproducing such pretraining requires access to sufficiently diverse,
    high-quality data and substantial computational resources, making it prohibitively
    expensive in many research and deployment settings.
    When developing new policy architectures, particularly efficient models~\citep{yu2025survey, guan2025efficient}, a natural goal is therefore to transfer capabilities from existing EFMs to target policies without repeating large-scale action pretraining.

    Models without action pretraining~\citep{lin2025evo,vlaadapter,starvla} can achieve competitive performance on standard simulation benchmarks such as LIBERO-Spatial and Object~\citep{libero}. Nevertheless, our pilot study in \Cref{fig:teaser} reveals larger performance gaps in more challenging settings, including long-horizon tasks in LIBERO-10, tasks under visual and robot-initial-state perturbations~\citep{libero_plus}, and real-world tasks. These observations motivate examining closed-loop capabilities beyond nominal action prediction. On a new downstream task, a student may produce reasonable actions at states well represented in the training demonstrations. Compounding execution errors and environmental stochasticity, however, can drive it into out-of-distribution states. This underscores the significance of transferring recovery behaviors from EFMs to help students recover from deviations, with the aim of improving closed-loop robustness.

    Policy distillation offers an intuitive route by transferring teacher
    knowledge within a task-specific domain~\citep{rusu2015policy, dong2025vita, zhang2025diffusion}.
    For closed-loop capabilities, effective distillation depends not only
    on \emph{what} actions are supervised, but also on \emph{where} supervision is
    provided. When distillation is restricted to a fixed offline dataset (\Cref{fig:teaser}(a)),
    supervision remains concentrated on demonstrated task execution, with limited coverage of states produced by the student's inferior predictions. This distribution mismatch
    between training and deployment leaves the student insufficiently prepared to
    handle execution deviations~\citep{yan2026human,hu2026rac}.

    Interactive imitation learning mitigates this mismatch by querying expert
    supervision at learner-induced states~\citep{dagger}. In on-policy distillation
    (OPD)~\citep{opd, agarwal2023gkd}, the student interacts with
    the environment while the teacher provides optimization targets at student-visited states (\Cref{fig:teaser}(b)). 
    These targets provide corrective supervision,
    but the queried teacher actions are not executed~\citep{zhong2026vla, wan2026adversarial}.
    Consequently, OPD does not observe the subsequent states reached by executing the teacher's corrections, which can offer additional training value in practical scenarios requiring multiple adjustments.
    Human-in-the-loop approaches such as HG-DAgger~\citep{kelly2019hg}
    (\Cref{fig:teaser}(c)) can collect such recovery trajectories but require continuous human
    monitoring and teleoperated takeover~\citep{murray2026flowdagger,xu2026compliant},
    placing substantial demands on human effort and limiting the scalability of data collection.

    To address these limitations, we introduce \method (\textbf{T}eacher-\textbf{I}n-the-\textbf{L}oop
    \textbf{D}istillation from \textbf{E}FMs), a policy distillation framework that
    transfers teacher knowledge over three complementary sources of experience:
    human demonstrations, student-induced rollouts, and teacher-executed recovery trajectories (\Cref{fig:teaser}(d)).
    To collect the required interaction data, we propose \ad, an autonomous teacher-in-the-loop data aggregation pipeline. An off-the-shelf robotic reward model monitors student execution, and an execution manager uses its feedback to trigger intervention. The EFM teacher then takes control and executes corrective actions toward task completion. The pipeline records both the preceding student rollout and the subsequent teacher recovery trajectory, making both sources available for training.
    By automating intervention decisions and corrective execution, \ad reduces reliance on expensive human oversight and avoids the overhead of teleoperated takeover, substantially enhancing the throughput of recovery-data collection.

    Given the aggregated data, \method applies supervision according to the sample source. On offline demonstrations and teacher recovery trajectories, we combine behavior cloning of recorded actions with distillation from teacher predictions, while student rollout states receive on-policy teacher supervision only. Importantly, all distillation is performed directly in the shared action space, without requiring alignment of internal representations, allowing the framework to readily accommodate heterogeneous teacher and student architectures.

    Our contributions are summarized as follows:
    \begin{enumerate}[noitemsep, topsep=0pt, leftmargin=*]
        \item We propose \method, a policy distillation framework that synergizes offline demonstrations, teacher supervision at student-visited states, and teacher-executed recovery experience for closed-loop capability transfer.

        \item We introduce \ad, an autonomous teacher-in-the-loop data aggregation pipeline that effectively collects student rollouts and teacher recovery trajectories through automatic task monitoring and teacher takeover, bypassing expensive human involvement.

        \item Extensive experiments across diverse student architectures on both real-world manipulation tasks and simulation benchmarks demonstrate that \method substantially improves student performance by up to $4.1\times$ over behavior cloning and enables effective cross-architecture transfer.
    \end{enumerate}

    \section{Related Work}

    \textbf{Embodied Foundation Models.} Recent advances in large-scale robot learning
    have given rise to Embodied Foundation Models (EFMs) that integrate visual perception,
    language understanding, and low-level control, showing impressive manipulation abilities across diverse environments~\citep{zhang2025pure,shao2025large,ma2024survey,xu2024survey}.
    Vision-Language-Action (VLA) models directly map multimodal observations and
    language instructions to robot actions~\citep{pi06,pi07,gr00t,xvla,fan2026any3d,faster}.
    An emerging alternative is the World Action Model (WAM), which jointly models
    future world states and robot actions, rather than relying solely on
    reactive observation-to-action prediction~\citep{ye2026world, lingbotva, yuan2026fast, wang2026openwam}.
    Despite their strong capabilities, training such foundation models from scratch
    requires substantial data and computational resources. Our work instead
    studies how these capabilities can be efficiently transferred to
    student policies, particularly for novel and lightweight architectures.

    \noindent
    \textbf{Policy Distillation.}
    Knowledge distillation (KD) transfers predictions or representations from a teacher
    to a student and is widely used for model compression and capability transfer~\citep{hinton2015distilling, compression, rusu2015policy}.
    Policy distillation extends this paradigm to robot learning~\citep{torne2024rialto, dong2025vita, shao2026x, zhang2025diffusion},
    with recent work leveraging foundation models as teachers~\citep{julg2025refined}.
    Existing approaches transfer knowledge through model-level alignment~\citep{jeon2026shallow, ye2025actdistill}
    or expert-generated data and supervision~\citep{shi2026offline, xiao2026self, xu2024rldg}.
    VLA-OPD~\citep{zhong2026vla} adopts on-policy distillation (OPD)~\citep{opd, agarwal2023gkd}
    to provide dense teacher supervision on student-generated trajectories. Beyond
    labeling student-visited states, \method executes teacher takeover when
    student stalls, collecting the subsequent recovery trajectories.
    These trajectories expose the student to states encountered during multi-step teacher
    recovery and provide direct supervision for returning from off-nominal
    states to successful task completion. Furthermore, VLA-OPD uses a 
    reverse-KL objective that simultaneously relies on teacher and student action-token probabilities, while
    \method operates in native action space without restrictions on architectures.

    \noindent
    \textbf{Interactive Imitation Learning.} Imitation learning suffers from distribution
    shift when the learned policy encounters states absent from expert demonstrations.
    Interactive imitation learning mitigates this issue by querying an expert on
    learner-induced states and aggregating the resulting data~\citep{dagger}.
    Subsequent methods reduce expert queries using predicted novice-expert discrepancy~\citep{hoque2021lazydagger},
    policy uncertainty~\citep{menda2019ensembledagger,zurbrugg2026vr},
    or estimates of state novelty and task risk~\citep{cronrath2018bagger, hoque2021thriftydagger}.
    However, these approaches generally depend on policy-specific techniques or
    continuous human oversight, and still request human-directed
    takeover on physical robots~\citep{kelly2019hg,murray2026flowdagger,xu2026compliant}. \ad instead combines a policy-independent task monitor with an EFM teacher to automatically
    detect intervention needs and execute corrective actions, substantially reducing
    the human-effort bottleneck.

    \section{Methodology}

    \subsection{Preliminaries}

    \noindent
    \textbf{Task Formulation.} We consider standard language-conditioned robotic
    manipulation with action-chunking policies~\citep{dp,act}. Given a
    state $\vs$ comprising RGB images and robot proprioception, together with a language
    instruction $\vl$, a policy $\pi$ predicts an action chunk:
    \begin{equation}
        \pi(\vs, \vl) = \mA = [\va_{1}, \dots, \va_{H}],
    \end{equation}
    where $H$ denotes the policy's prediction horizon and $\va_{h}$ is the $h$-th
    action in the chunk. During closed-loop execution, the policy is repeatedly queried
    using updated observations.

    \noindent
    \textbf{DAgger.} Dataset Aggregation (DAgger)~\citep{dagger} is an interactive
    imitation learning algorithm that addresses the distribution shift between offline
    demonstrations $\mathcal{D}_{\mathrm{Demo}}$ and learner rollouts $\mathcal{D}
    _{\mathrm{Roll}}$ by collecting expert action labels at learner-induced states.
    In the representative human-in-the-loop variants~\cite{kelly2019hg}, the human expert monitors the
    task progress and takes control when the learner's behavior requires correction
    by providing recovery actions $\va^{\star}$ at the encountered state $\vs$.
    The resulting labeled samples $(\vs, \vl, \va^{\star})$ are added to an
    aggregated dataset $\mathcal{D}_{\mathrm{Rec}}$, which is used to update the
    learner through behavior cloning.

    \subsection{Overview}

    \method transfers closed-loop capabilities from a teacher policy
    $\pi_\mathrm{T}$ to a student policy $\pi_\mathrm{S}$. We obtain the
    teacher by finetuning a pretrained EFM on the human demonstration
    dataset $\mathcal{D}_{\mathrm{Demo}}$ for the downstream embodiment and task,
    and keep it frozen throughout the framework. The student is initialized
    with a pretrained backbone (\eg, a VLM for a VLA model), receives no
    robot action pretraining, and is initially trained on
    $\mathcal{D}_{\mathrm{Demo}}$ using behavior cloning.
    \method then proceeds in two stages:
    (1) During \ad data collection, the student executes the task while
    a task monitor estimates progress and an execution manager determines
    when the teacher should intervene. Upon intervention, the teacher
    takes control to perform recovery. Only episodes with confirmed
    recovery are retained, each comprising the pre-intervention
    student rollout and the subsequent teacher recovery trajectory.
    (2) Action-space distillation is then applied on three complementary
    data sources: offline demonstrations, student rollouts, and teacher
    recovery trajectories.

    \subsection{AutoDAgger}
    \label{sec:ad}

    \begin{figure*}
        \centering
        \includegraphics[width=0.9\linewidth]{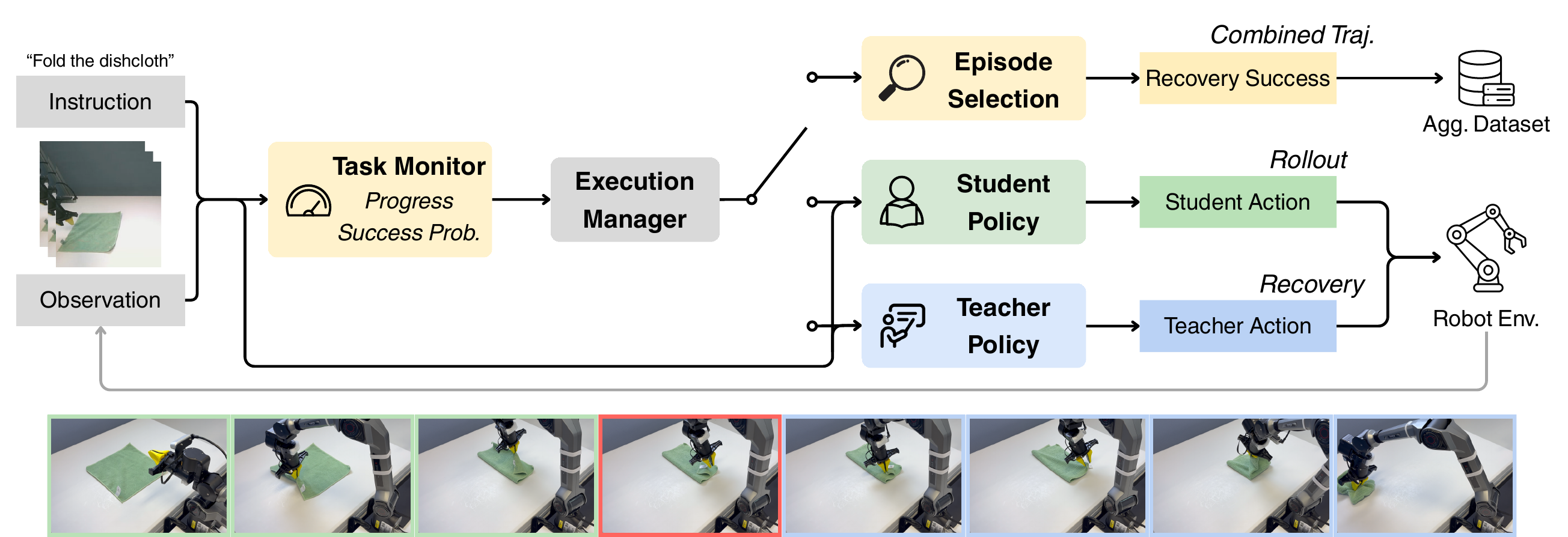}
        \vspace{1mm}
        \caption{\textbf{AutoDAgger pipeline.} A reward-model-based task monitor
        estimates task progress and success probability. The execution manager
        uses these signals to coordinate policy execution, trigger teacher intervention
        upon detected student failure, and select successful recovery episodes
        for data aggregation. Pre-intervention student rollout and successful teacher
        recovery are recorded without human monitoring or takeover. \textbf{Bottom:}
        A real-world Fold Dishcloth example; green/blue denote student/teacher
        execution, while red marks the detected failure and intervention point.}
        \label{fig:autodagger}
        \vspace{-5mm}
    \end{figure*}

    To reduce human supervision and collect informative correction data more efficiently,
    we propose \ad, an autonomous teacher-in-the-loop pipeline that monitors student
    execution, detects task-level failures, triggers teacher takeover, and
    retains successfully recovered trajectories for subsequent policy
    improvement.

    \noindent
    \textbf{Task Monitor.} We employ a general robotic reward model $R$ as
    a policy-independent task monitor. We index monitor updates by
    $t=1,2,\ldots$ and denote the timestamp of the $t$-th update by $\tau_{t}$. For the $t$-th monitor update, we use $\vo_{1:K_t}$,
    a sequence of $K_t$ observations sampled
    at evenly spaced frame indices from the visual history,
    including the latest observation. Given these observations and the language instruction
    $\vl$, the monitor predicts a task progress score and a success probability
    for each sampled observation:
    \begin{equation}
        \left(\hat{\vp}_{1:K_t},\hat{\vq}_{1:K_t}\right) = R\left(\vo_{1:K_t},\vl
        \right). \label{eq:robometer_predictions}
    \end{equation}
    The output sequences describe estimations over the sampled history. We use
    the final entry of each sequence, corresponding to the latest observation,
    as the current task progress $p_{t}$ and success probability $q_{t}$, respectively.

    \noindent
    \textbf{Execution Manager.} The execution manager coordinates policy
    execution, teacher intervention, and data selection based on information
    from the task monitor. Each episode begins with the student policy, while
    detected failures trigger teacher takeover. To identify failures, we check
    for sustained progress stagnation when the task is not currently predicted
    to be successful. Specifically, to avoid interpreting small fluctuations in
    the predicted progress $p_{t}$ as task advancement, we maintain a reference progress
    value $\bar{p}_{t}$, updated only when $p_{t}$ exceeds the previous
    reference by a margin $\epsilon_{p}$. The meaningful-progress indicator $g_{t}$
    and reference update are defined as
    \begin{equation}
        g_{t}= \mathbb{I}\!\left[p_{t}>\bar{p}_{t-1}+\epsilon_{p}\right], \quad \bar
        {p}_{t}=
        \begin{cases}
            p_{t},         & g_{t}=1, \\
            \bar{p}_{t-1}, & g_{t}=0.
        \end{cases}
        \label{eq:progress_update}
    \end{equation}
    The execution manager interprets $q_{t}>\theta_{s}$ as an instantaneous prediction
    of task success. We regard either meaningful progress or such a success
    prediction as evidence of valid execution and track the timestamp of the
    most recent evidence:
    \begin{equation}
        \tau_{t}^{\mathrm{valid}}=
        \begin{cases}
            \tau_{t},                    & g_{t}=1 \;\lor\; q_{t}>\theta_{s}, \\
            \tau_{t-1}^{\mathrm{valid}}, & \text{otherwise}.
        \end{cases}
        \label{eq:valid_time_update}
    \end{equation}
    A failure is detected when the elapsed time since the last valid execution reaches
    $T_{\mathrm{fail}}$ and the current success probability does not exceed $\theta
    _{s}$:
    \begin{equation}
        \Delta_{t}= \tau_{t}-\tau_{t}^{\mathrm{valid}}, \quad F_{t}= \mathbb{I}\!
        \left[ \Delta_{t}\geq T_{\mathrm{fail}}\;\land\; q_{t}\leq\theta_{s}\right
        ]. \label{eq:intervention_decision}
    \end{equation}
    When $F_{t}=1$ during student execution, the manager immediately halts the student
    and transfers control to the teacher policy. The same monitoring mechanism
    is applied during teacher execution. Task completion is confirmed when consecutive
    $q_{t}$ predictions remain above $\theta_{s}$ for at least $T_{\mathrm{succ}}$,
    with any prediction at or below the threshold resetting the success timer. If
    teacher recovery fails, the episode is discarded. Otherwise, the student
    rollout and teacher recovery segment are retained as a combined trajectory
    and aggregated for subsequent policy training.

    \subsection{Distillation Objective}
    \label{sec:distillation} Given the interaction data collected by \ad, \method trains
    the student from three sources of experience: offline
    demonstrations $\mathcal{D}_{\mathrm{Demo}}$, states visited during student rollouts
    $\mathcal{D}_{\mathrm{Roll}}$, and teacher-executed recovery trajectories
    $\mathcal{D}_{\mathrm{Rec}}$. For a state-instruction pair $(\vs,\vl)$,
    we perform distillation directly in the action space:
    \begin{equation}
        \mathcal{L}_{\mathrm{KD}}(\vs,\vl) = \ell_\mathrm{S}\left( \vs, \vl; \operatorname{sg}
        \! \left[\mA_\mathrm{T}\right] \right), \label{eq:kd_loss}
    \end{equation}
    where $\operatorname{sg}[\cdot]$ stops gradients through the frozen teacher
    and $\ell_\mathrm{S}$ denotes the original action-generation objective used by the
    student, instantiated with the teacher-predicted action chunk $\mA_\mathrm{T}=\pi_\mathrm{T}
    (\vs,\vl)$ as the target. Since supervision is imposed only on action
    outputs, \method does not require the teacher and student to share similar architectures
    or intermediate representations.

    \begin{figure*}[t]
        \centering
        \begin{minipage}[t]{0.67\linewidth}
            \vspace{0pt}
            \centering
            \includegraphics[width=\linewidth]{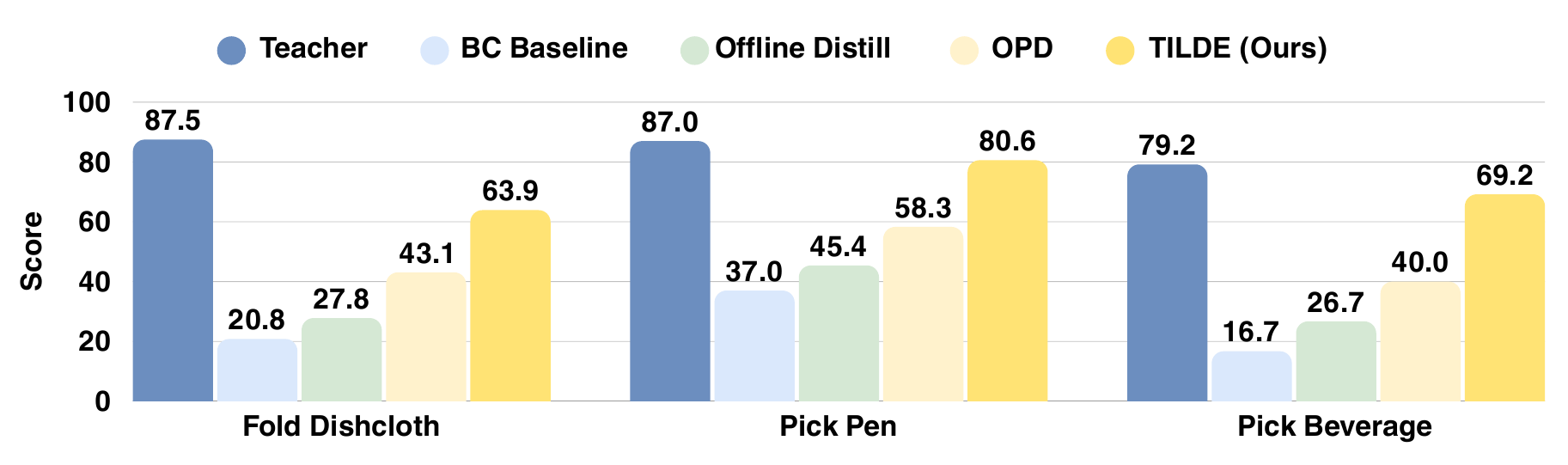}
            \caption{\textbf{Comparison of real-world task performance.} \method{}
            is compared with behavior cloning (BC), offline distillation, and on-policy
            distillation (OPD), with the teacher policy serving as a reference.
            \method{} outperforms all student baselines on all three tasks.}
            \label{fig:main_results}
        \end{minipage}%
        \hfill%
        \begin{minipage}[t]{0.3\linewidth}
            \vspace{0pt}
            \centering
            \captionof{table}{\textbf{Standardized failure recovery evaluation.} ``Recovery SR'' is measured from the same set of 12 predefined error states on Fold Dishcloth.}
            \vspace{2mm}
            \label{tab:recover} \small
            \setlength{\tabcolsep}{12pt}
            \begin{tabular}{cc}
                \toprule Method  & Recovery SR   \\
                \midrule \gray{Teacher} & \gray{9/12}          \\
                BC Baseline      & 4/12          \\
                Offline Distill  & 5/12          \\
                OPD              & 5/12          \\
                \method{} (Ours) & \textbf{8/12} \\
                \bottomrule
            \end{tabular}
        \end{minipage}
        \vspace{-4mm}
    \end{figure*}

    For samples from either offline demonstrations or recovery trajectories, the
    action actually executed in the environment is available. Let $\mA_\mathrm{E}$ denote
    this recorded action chunk. We therefore combine behavior cloning from the
    executed action with distillation from the teacher prediction:
    \begin{equation}
        \begin{aligned}
            \mathcal{L}_{\mathrm{Exec}}(\vs,\vl,\mA_\mathrm{E}) & = \ell_\mathrm{S}(\vs,\vl;\mA_\mathrm{E}) + \lambda_{\mathrm{KD}}\mathcal{L}_{\mathrm{KD}}(\vs,\vl), \\
            (\vs,\vl ,\mA_{E})                           & \in \mathcal{D}_{\mathrm{Demo}}\cup \mathcal{D}_{\mathrm{Rec}}, \label{eq:exec_loss}
        \end{aligned}
    \end{equation}
    where $\lambda_{\mathrm{KD}}$ is the weight of distillation loss.
    In contrast, as samples from $\mathcal{D}_{\mathrm{Roll}}$ contain action generated
    by the student, we apply only the distillation objective:
    \begin{equation}
        \mathcal{L}_{\mathrm{Roll}}(\vs,\vl) = \lambda_{\mathrm{KD}}\mathcal{L}_{\mathrm{KD}}
        (\vs,\vl), \quad (\vs,\vl)\in\mathcal{D}_{\mathrm{Roll}}. \label{eq:rollout_loss}
    \end{equation}

    Overall, the training loss is selected according to the source of each sample.
    This formulation exposes the student to three complementary
    distributions. Offline demonstrations provide supervision on in-domain expert
    behavior, rollouts provide teacher corrections at student-induced
    states, and recovery trajectories supply the subsequent observations covered
    by executing the teacher's corrections.

    \section{Experiments}
    We evaluate \method across policy architectures, real-world
    tasks, and simulation benchmarks. In \Cref{sec:exp_main,sec:exp_ana}, we
    pair a $\pi_{0.5}$ teacher with a student of the same
    architecture but without action pretraining ($\pi_{0.5}$ w/o PT).
    This setting controls for architectural differences to isolate
    the effectiveness of our distillation framework. In \Cref{sec:exp_cross},
    we evaluate cross-architecture transfer using X-VLA
    and SmolVLA as the students, both without action pretraining. Finally,
    \Cref{sec:exp_exec} analyzes the failure detection and intervention
    timing of the execution manager in \ad, and the contribution of collected data to distillation.

    \subsection{Experimental Setup}
    \label{sec:exp_setup}

    \noindent
    \textbf{Tasks and Evaluation.}
    We conduct real-world experiments using 7-DoF AgileX PiperX robot arms
    on three manipulation tasks:
    (1) ``Fold Dishcloth'': folding a dishcloth in half twice and placing
    it in a designated front-left area,
    testing deformable object manipulation;
    (2) ``Pick Pen'': picking up one to three pens and placing them in
    a cup, testing precise manipulation across varying object locations;
    and (3) ``Pick Beverage'': picking up a beverage and placing it in
    a paper box, testing generalization across objects and locations.
    Our offline dataset comprises 200 human demonstration episodes for
    Fold Dishcloth and 100 episodes each for Pick Pen and Pick Beverage.
    
    We evaluate all policies on the same predefined test cases with
    fixed object types, positions, and orientations. We conduct 12 trials for Fold Dishcloth, 27 for Pick Pen, and 30 for Pick Beverage.
    We report fine-grained task scores averaged across sub-steps and evaluation trials to enable
    a more precise assessment of policy performance.
    Detailed scoring criteria are provided in the supplementary material.
    
    We additionally evaluate on the LIBERO simulation
    benchmark~\citep{libero}, which comprises four suites:
    Spatial, Object, Goal, and 10.
    Each suite contains 10 tasks, with each task evaluated over
    50 rollouts. We report the average task success rate for each suite.

    \noindent
    \textbf{Implementation Details.}
    For all experiments, we obtain the teacher by fine-tuning the pretrained
    $\pi_{0.5}$ model~\citep{pi05} on the corresponding offline
    dataset. 
    Students receive no action pretraining: their action experts are
    trained from scratch, while the backbones are initialized from
    pretrained VLMs. Specifically, PaliGemma~\citep{paligemma} is utilized
    for $\pi_{0.5}$ w/o PT, Florence-2-Large~\citep{xiao2024florence} for X-VLA w/o PT,
    and SmolVLM2-500M~\citep{marafioti2025smolvlm} for SmolVLA w/o PT.
    The task monitor in \ad uses the off-the-shelf robotic reward model
    Robometer~\citep{liang2026robometer} without task-specific fine-tuning.
    Further implementation details are provided in the supplementary material.

    \begin{figure*}[t]
        \centering
        \includegraphics[width=\linewidth]{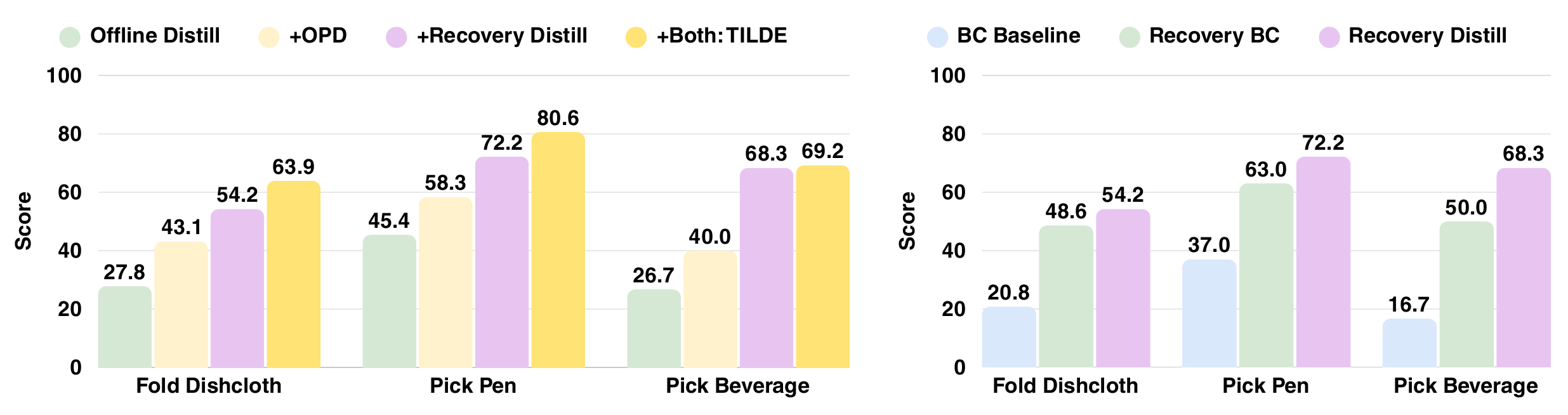}
        \caption{\textbf{Ablation study of training data and supervision. } \textbf{Left:}
        Starting from offline distillation, we add teacher supervision on student-rollout
        data (+OPD), supervision along teacher recovery trajectories (+Recovery Distill),
        or both (\method). Combining both sources yields the highest scores across
        all three tasks. \textbf{Right:} Recovery BC augments demonstration BC
        with imitation of recorded recovery actions. Recovery Distill additionally
        incorporates distillation on both demonstration and recovery samples,
        further improving performance.}
        \label{fig:ablation}
        \vspace{-4mm}
    \end{figure*}

    \noindent
    \textbf{Baselines and Training Protocol.} 
    We compare \method with three training paradigms:
    behavior cloning (BC)~\citep{bain1995framework,pomerleau1988alvinn},
    offline distillation, and on-policy distillation (OPD).
    BC learns from recorded human actions. Offline distillation additionally
    uses teacher action targets at demonstration states, while OPD further
    incorporates teacher supervision at states visited during student rollouts.
    Unless otherwise stated, comparisons regarding interaction data
    use a shared pool of
    \ad data: 30 episodes per real-world task and 50 episodes per LIBERO suite.

    \subsection{Main Results}
    \label{sec:exp_main}

    \Cref{fig:main_results} shows that offline distillation yields only modest
    improvements over BC, as teacher supervision remains confined to the demonstration
    distribution. OPD yields further gains by mitigating distribution mismatch
    through extending teacher supervision to student-rollout states. \method achieves
    scores of 63.9, 80.6, and 69.2 on Fold Dishcloth, Pick Pen, and Pick Beverage,
    respectively, outperforming all evaluated baselines on each task. These
    scores are approximately $3.1\times$, $2.2\times$, and $4.1\times$ those
    of BC baseline, substantially narrowing the observed student-teacher
    performance gap. These consistent improvements support the value of learning
    from recovery data in addition to teacher action targets at student's own distributions.

    \begin{table}[t]
        \centering
        \caption{\textbf{Scaling with \ad data.} Zero episodes
        corresponds to offline distillation on human demonstrations alone.
        }
        \setlength{\tabcolsep}{7pt}
        \small
        \begin{tabular}{c|ccc}
            \toprule \#Episodes & Fold Dishcloth & Pick Pen  & Pick Beverage \\
            \midrule 0          & 27.8  & 45.4 & 26.7    \\
            30                  & 63.9  & 80.6 & 69.2   \\
            60                  & 65.3  & 81.5 & 71.7    \\
            90                  & 77.8  & 86.1 & 75.0    \\
            \bottomrule
        \end{tabular}
        \label{tab:data_scaling}
        \vspace{-3mm}
    \end{table}

    \begin{table*}
        [t]
        \centering
        \caption{\textbf{Comparison of additional human
        demonstrations and \ad data.}
        We augment the original dataset with 30 episodes from either source. ``Distill'' indicates the use of teacher action
        targets, and \#Episodes is the total number of training episodes.
        Gains are absolute score improvements over Human Demo under the
        same distillation setting. \ad data yield substantially larger gains across
        all settings.}
        \setlength{\tabcolsep}{10.5pt}
        \small
        \begin{tabular}{lc|cccccc}
            \toprule \multirow{2}{*}{Data Source} & \multirow{2}{*}{Distill} & \multicolumn{2}{c}{Fold Dishcloth} & \multicolumn{2}{c}{Pick Pen} & \multicolumn{2}{c}{Pick Beverage} \\
                                                  &                          & \#Episodes                         & Score                        & \#Episodes                       & Score                       & \#Episodes & Score                       \\
            \midrule Human Demo                   & \xmark                   & 200                                & 20.8                         & 100                              & 37.0                        & 100        & 16.7                        \\
            \quad +Extra Demo                     & \xmark                   & 230                                & 27.8\,\gain{+7.0}            & 130                              & 45.4\,\gain{+8.4}           & 130        & 44.2\,\gain{+27.5}          \\
            \quad +\ad                            & \xmark                   & 230                                & \textbf{48.6}\,\gain{+27.8}  & 130                              & \textbf{63.0}\,\gain{+26.0} & 130        & \textbf{50.0}\,\gain{+33.3} \\
            \midrule Human Demo                   & \cmark                   & 200                                & 27.8                         & 100                              & 45.4                        & 100        & 26.7                        \\
            \quad +Extra Demo                     & \cmark                   & 230                                & 29.2\,\gain{+1.4}            & 130                              & 54.6\,\gain{+9.2}           & 130        & 48.3\,\gain{+21.6}          \\
            \quad +\ad                            & \cmark                   & 230                                & \textbf{63.9}\,\gain{+36.1}  & 130                              & \textbf{80.6}\,\gain{+35.2} & 130        & \textbf{69.2}\,\gain{+42.5} \\
            \bottomrule
        \end{tabular}
        \label{tab:data_source}
        \vspace{-2mm}
    \end{table*}

    \begin{table*}
        [t]
        \centering
        \caption{ \textbf{Comparison of HG-DAgger and AutoDAgger in policy
        performance, recovery success, and data collection efficiency.}
        $T_{\mathrm{Rec}}$ denotes the average time required for the human operator or
        teacher policy to achieve successful recovery after takeover. Rec. SR
        is the fraction of interventions that result in successful recovery and
        task completion. $T_{\mathrm{Avg}}$ denotes the average end-to-end collection
        time per valid episode, including time spent on failed attempts. Collection metrics
        are evaluated over 30 additional trials. Parenthesized values indicate
        the speedups achieved by \ad over HG-DAgger. }
        \label{tab:human_vs_auto}
        \setlength{\tabcolsep}{2.9pt}
        \small
        \begin{tabular}{cc|cccccccccccc}
            \toprule \multirow{2}{*}{Method}      & \multirow{2}{*}{Distill} & \multicolumn{4}{c}{Fold Dishcloth} & \multicolumn{4}{c}{Pick Pen} & \multicolumn{4}{c}{Pick Beverage} \\
                                                  &                          & Score                              & $T_{\mathrm{Rec}}$           & Rec. SR (\%)                          & $T_{\mathrm{Avg}}$     & Score & $T_{\mathrm{Rec}}$     & Rec. SR (\%)                 & $T_{\mathrm{Avg}}$     & Score & $T_{\mathrm{Rec}}$     & Rec. SR (\%)                 & $T_{\mathrm{Avg}}$     \\
            \midrule

\multirow{2}{*}{HG-DAgger}  & \xmark                   & 34.7                               & \multirow{2}{*}{54.4s}       & \multirow{2}{*}{76.7}          & \multirow{2}{*}{95.8s} & 75.9  & \multirow{2}{*}{29.4s} & \multirow{2}{*}{76.7} & \multirow{2}{*}{46.6s} & 60.0  & \multirow{2}{*}{32.2s} & \multirow{2}{*}{90.0} & \multirow{2}{*}{39.3s} \\
                                                  & \cmark                   & 65.3                               &                              &                                  &                        & 80.6  &                        &                         &                        & 63.3  &                        &                         &                        \\
            \midrule

\multirow{2}{*}{AutoDAgger} & \xmark                   & 48.6                               & 21.5s                        & \multirow{2}{*}{73.3}          & 48.8s                  & 63.0  & 13.2s                  & \multirow{2}{*}{66.7} & 30.9s                  & 50.0  & 12.4s                  & \multirow{2}{*}{90.0} & 20.3s                  \\
                                                  & \cmark                   & 63.9                               & \gain{2.5$\times$}           &                                  & \gain{2.0$\times$}     & 80.6  & \gain{2.2$\times$}     &                         & \gain{1.5$\times$}     & 69.2  & \gain{2.6$\times$}     &                         & \gain{1.9$\times$}     \\
            \bottomrule
        \end{tabular}
        \vspace{-4mm}
    \end{table*}

    \subsection{In-depth Analysis}
    \label{sec:exp_ana}

    \noindent
    \textbf{Ablation Study.}
    We first examine the contributions of supervision at student-rollout
    states and along teacher-executed recovery trajectories, using offline
    distillation as a common baseline. As shown in \Cref{fig:ablation}
    (left), adding either source improves performance on all three tasks,
    while combining both produces the highest scores. Each source therefore
    provides gains beyond those obtained from the other alone, supporting
    their complementary roles in transferring teacher capabilities.
    
    \Cref{fig:ablation} (right) further analyzes the supervision used
    on demonstration and recovery data. Recovery BC augments conventional
    BC with imitation of recorded teacher recovery actions, improving
    downstream performance. Recovery Distill additionally applies
    knowledge distillation to both demonstration and recovery samples, achieving further advantages.

    \noindent
    \textbf{Recovery from Failures.} 
    To directly assess recovery capabilities, we construct a standardized recovery-state evaluation
    with a common set of 12 distinct error states on Fold Dishcloth.
    Each state deviates from nominal execution and requires corrective
    behavior to complete the task. As shown in \Cref{tab:recover},
    BC succeeds from 4 of the 12 states, while offline distillation and
    OPD improve this result slightly to 5/12. \method succeeds from 8 states, approaching
    the pretrained teacher's 9. These results support the value of
    learning from the subsequent states generated by executing teacher
    corrections, which expose
    the student to the states and actions involved in correction behavior.

    \noindent
    \textbf{Scaling with \ad Data.}
    \Cref{tab:data_scaling} examines how performance scales with the
    amount of \ad data.
    Starting from offline-distillation scores of 27.8, 45.4, and 26.7
    on Fold Dishcloth, Pick Pen, and Pick Beverage, respectively,
    adding 30 \ad episodes per task raises these scores to 63.9,
    80.6, and 69.2. Increasing the data to 60 and 90 episodes yields
    further improvements on every task. With 90 episodes, the student
    reaches scores of 77.8, 86.1, and 75.0, approaching the teacher's
    87.5, 87.0, and 79.2, respectively. These results demonstrate
    consistent benefits from continued scaling of interactive recovery data.

    \noindent
    \textbf{Quality and Efficiency of \ad Data.} \Cref{tab:data_source} compares
    augmenting the original dataset with the same number of additional human
    teleoperation episodes or \ad episodes per task. Under BC, \ad data yield
    scores of 48.6, 63.0, 50.0 on Fold Dishcloth, Pick Pen, and Pick Beverage, compared
    with 27.8, 45.4, and 44.2 using additional human demonstrations. With
    distillation enabled, the \ad configuration achieves scores of 63.9, 80.6, and
    69.2 across the three tasks, again substantially outperforming the additional-demonstration
    configuration. These results suggest that \ad data provide more effective supervision
    than conventional expert demonstrations, which typically consist of
    trajectories that directly lead to task success. In contrast, \ad provides
    corrective supervision at student-induced states that are rarely encountered
    during standard demonstration collection, while also enabling on-policy distillation
    to more effectively leverage the teacher's knowledge.

    We further compare \ad with HG-DAgger~\citep{kelly2019hg}, a representative human-in-the-loop
    data aggregation method, in terms of downstream policy performance, recovery
    success, and data collection efficiency. As shown in \Cref{tab:human_vs_auto},
    after distillation, \ad achieves competitive performance across all three tasks:
    it closely matches HG-DAgger on Fold Dishcloth (63.9 \textit{vs.} 65.3),
    achieves the same score on Pick Pen, and obtains a higher score on Pick Beverage
    (69.2 \textit{vs.} 63.3).

    \ad substantially reduces recovery time ($T_{\mathrm{Rec}}$), achieving a speedup
    of over 2$\times$ on each task. This improvement is facilitated by direct teacher-policy
    takeover: once an intervention is triggered, the teacher generates and executes
    corrective actions from the current state without manual control. \ad also
    automates progress monitoring and intervention decisions, eliminating the need
    for a human operator to continuously supervise execution and manually initiate
    takeover.

    To quantify end-to-end collection efficiency, we additionally report
    $T_{\mathrm{Avg}}$, the average time required to collect a valid episode,
    including time spent on failed attempts. Although \ad may have lower recovery
    success rates than HG-DAgger, it consistently reduces $T_{\mathrm{Avg}}$, yielding
    collection speedups of 1.5--2.0$\times$. These results show that the
    efficiency advantage persists even after accounting for student rollout time
    and unsuccessful attempts. By combining higher collection throughput with automated
    monitoring and intervention, \ad offers a more scalable alternative to human-in-the-loop
    data aggregation.

    Higher collection throughput also provides an opportunity to close the small
    performance gap through additional autonomous data collection. On Fold Dishcloth,
    increasing the \ad dataset to 60 episodes raises the score to 65.3 (\Cref{tab:data_scaling}),
    matching HG-DAgger with 30 episodes. Combining with the 2$\times$ end-to-end collection
    speedup, this result demonstrates that \ad can collect twice as many valid
    episodes and achieve comparable downstream performance within a similar time
    budget.

    \subsection{Cross-architecture Distillation}
    \label{sec:exp_cross}

    We investigate whether \method can transfer capabilities across
    heterogeneous policy architectures. By aligning teacher and student
    outputs solely in the action space, our framework does not require alignment
    of intermediate representations, allowing the two policies to differ
    in architecture or action-generation mechanism.
    
    On the real-world tasks, we distill the $\pi_{0.5}$ teacher into
    X-VLA, a compact student policy with only 0.88B parameters.
    \Cref{tab:cross_xvla} illustrates that \method significantly improves performance score by
    30.6, 36.1, and 32.5 on the three tasks,
    respectively. The distilled student also outperforms the X-VLA model
    pretrained on over 200K episodes across all three tasks, demonstrating
    effective capability transfer to a different policy architecture.
    
    We further evaluate on LIBERO using
    SmolVLA~\citep{smolvla}, a 0.45B-parameter student. As shown in \Cref{tab:cross_smolvla}, \method outperforms
    both the student baseline and offline distillation on all four suites,
    achieving an average success rate of 89.6\%, compared with 81.2\%
    and 83.5\%, respectively. The largest improvement over the
    baseline occurs on the long-horizon LIBERO-10 suite, where success
    increases from 64.6\% to 79.6\%, consistent with our motivation to transfer capabilities needed in more
    challenging settings. These results demonstrate the
    applicability of \method across policy architectures and support
    the use of pretrained EFMs as reusable teachers for new student
    architectures.

    \begin{table}[t]
        \centering
        \caption{\textbf{Cross-architecture distillation on real-world tasks
        with X-VLA as the student.} Starting from X-VLA without action pretraining
        (w/o PT), \method improves performance across all three tasks and
        outperforms the action-pretrained variant (w/ PT). Gains are 
        absolute score improvements over X-VLA w/o PT.}
        \setlength{\tabcolsep}{2pt}
        \small
        \begin{tabular}{c|cccc}
            \toprule Method                       & Params      & Dishcloth                   & Pen                         & Beverage                    \\
            \midrule \gray{$\pi_{0.5}$ (Teacher)} & \gray{3.4B} & \gray{87.5}                 & \gray{87.0}                 & \gray{79.2}                 \\
            X-VLA w/ PT                           & 0.88B       & 19.4                        & 45.4                        & 26.7                        \\
            X-VLA w/o PT                          & 0.88B       & 9.7                         & 41.7                        & 10.0                        \\
            \method (Ours)                        & 0.88B       & \textbf{40.3}\,\gain{+30.6} & \textbf{77.8}\,\gain{+36.1} & \textbf{42.5}\,\gain{+32.5} \\
            \bottomrule
        \end{tabular}
        \label{tab:cross_xvla}
    \end{table}

    \begin{table}[t]
        \centering
        \caption{\textbf{Cross-architecture distillation on LIBERO with SmolVLA
        as the student.} \method
        outperforms both the student baseline and offline distillation on all
        four suites.}
        \setlength{\tabcolsep}{3pt}
        \small
        \begin{tabular}{c|cccccc}
            \toprule Method                       & Params      & Spatial       & Object        & Goal          & 10            & Avg.          \\
            \midrule \gray{$\pi_{0.5}$ (Teacher)} & \gray{3.4B} & \gray{98.8}   & \gray{98.2}   & \gray{98.0}   & \gray{92.4}   & \gray{96.9}   \\
            SmolVLA w/o PT                              & 0.45B       & 81.4          & 94.4          & 84.2          & 64.6          & 81.2          \\
            Offline Distill                       & 0.45B       & 85.2          & 96.4          & 88.4          & 63.8          & 83.5          \\
            \method (Ours)                        & 0.45B       & \textbf{88.0} & \textbf{96.6} & \textbf{94.0} & \textbf{79.6} & \textbf{89.6} \\
            \bottomrule
        \end{tabular}
        \label{tab:cross_smolvla}
        \vspace{-4mm}
    \end{table}

    \subsection{Validation of Execution Manager}
    \label{sec:exp_exec}
 
    The execution manager in \ad determines whether and when the teacher should
    take over from the student. We evaluate both its ability to identify
    failures and the effect of intervention timing on recovery data collection and
    downstream distillation. We conduct these experiments on LIBERO-10,
    which provides a controlled and reproducible environment.
    
    \noindent
    \textbf{Failure Detection.}
    Using a fixed SmolVLA checkpoint, we collect 200 episodes and compare
    the execution manager's failure predictions with the ground-truth
    outcomes provided by the simulator. Treating failure as the positive
    class, the manager achieves 91.2\% precision, 99.2\% recall,
    and an F1 score of 95.0\% (\Cref{tab:intervention_acc}).
    Of the 136 predicted failures, 12 correspond to successful episodes
    and would therefore trigger unnecessary interventions. Conversely,
    only 1 of the 125 failed episodes is missed. These results indicate
    that the manager identifies nearly all failures while keeping
    unnecessary interventions relatively infrequent.
    
    \noindent
    \textbf{Intervention Timing.}
    We further investigate how intervention timing affects recovery
    data collection and downstream distillation by comparing the \ad
    trigger with earlier, later, fixed-step, and random interventions.
    As shown in \Cref{tab:intervention_timing}, earlier takeover
    generally improves task completion during collection by intervening
    before failures fully develop. In contrast, delayed intervention
    makes recovery more difficult and may allow errors to escalate
    into irrecoverable states, resulting in inferior collection efficiency.
    
    However, higher collection success does not necessarily translate
    into better distillation performance. For a fair comparison, we
    sample the same amount of data for distillation across all strategies.
    Earlier takeover offers limited additional supervision, as the
    resulting trajectories more closely resemble those in the
    offline demonstration dataset. \ad achieves the best downstream
    performance, highlighting the
    importance of when corrective supervision is collected, beyond
    whether collection episodes succeed. These results support the
    execution manager's ability to identify failures and trigger
    timely interventions that yield useful recovery data for distillation.

    \begin{table}[t]
        \centering
        \caption{\textbf{Failure detection by the execution manager on
        LIBERO-10.} Predictions are evaluated
        against simulator ground truth, treating failure as
        the positive class. Unnecessary interventions are false positives
        among all predicted failures; missed failures are false negatives
        among all actual failures.}
        \small
        \setlength{\tabcolsep}{8pt}
        \begin{tabular}{ccccc}
        \toprule
        \multirow{2}{*}{Precision} & \multirow{2}{*}{Recall} & \multirow{2}{*}{F1} & Unnecessary & Missed  \\
        & & & interventions & failures \\
        \midrule
        91.2\% & 99.2\% & 95.0\% & 12/136 & 1/125  \\
        \bottomrule
        \end{tabular}
        \label{tab:intervention_acc}
        \vspace{-2mm}
    \end{table}

    \begin{table}[t]
        \centering
    \caption{\textbf{Effect of intervention timing on LIBERO-10.}
    $s_\mathrm{AD}$ denotes the time step selected by \ad. Later, fixed, or random interventions may occur after an episode has already
    terminated, reducing the number of episodes with intervention.}
        \small
        \setlength{\tabcolsep}{9pt}
        \begin{tabular}{c|cc}
        \toprule
        Intervention Step & Recovery SR (\%) & Distill SR (\%) \\
        \midrule
        \gray{Offline Distill} & \gray{--} & \gray{71.8} \\
        $s_\mathrm{AD}-50$ & 106/136 (77.9) & 75.8 \\
        $s_\mathrm{AD}-20$ & 74/136 (54.4) & 79.0 \\
        $s_\mathrm{AD}$ & 75/136 (55.1) & \textbf{80.8} \\
        $s_\mathrm{AD}+20$ & 69/136 (50.7) & 79.2 \\
        $s_\mathrm{AD}+50$ & 58/135 (43.0) & 76.2 \\
        \midrule
        Fixed 50 & 166/200 (83.0) & 77.8 \\
        Fixed 100 & 137/199 (68.8) & 73.2 \\
        Fixed 200 & 59/141 (41.8) & 73.8 \\
        Random (0, 520) & 68/153 (44.4) & 77.2 \\
        \bottomrule
        \end{tabular}
        \label{tab:intervention_timing}
        \vspace{-3mm}
    \end{table}

    \section{Conclusion}

    In this work, we study how closed-loop capabilities acquired through large-scale action pretraining
    can be transferred from EFMs to student policies without repeating
    large-scale pretraining.
    We presented \method, a framework combining offline demonstrations,
    teacher supervision at student-visited states, and teacher-executed
    recovery trajectories to extend supervision to the states and
    actions encountered during recovery from execution errors. \ad supports
    this process through automated task monitoring and teacher takeover,
    reducing the human effort required to collect recovery experience.
    Real-world experiments demonstrate consistent improvements over behavior
    cloning, offline distillation, and on-policy distillation.
    Cross-architecture experiments with X-VLA and SmolVLA further validate
    the flexibility of action-space distillation. These results support the use of EFMs as reusable
    teachers for developing robust student policies.

    { \small \bibliographystyle{ieeenat_fullname} \bibliography{main} }

    \clearpage

\twocolumn[{\begin{center}
    {\LARGE Appendix}
\end{center}}]

We provide the following contents in the appendix:
\begin{itemize}[noitemsep, topsep=0pt, leftmargin=*]
    \item \Cref{supp:related} discusses additional related work.
    \item \Cref{supp:imple} provides implementation details of \ad, real-world
        experiments, simulation benchmarks, data collection, training configurations, and robot deployment.

    \item \Cref{supp:exp} presents additional experimental results, including
        execution manager evaluation, statistics, and visualizations of the \ad
        collection process.

    \item \Cref{supp:limitation} discusses limitations and potential directions for future work.
\end{itemize}

\section{Additional Related Work}
\label{supp:related}

We provide additional comparisons with closely related approaches,
focusing on their training objectives, teacher policies, and
intervention mechanisms.
\begin{itemize}
\item Refined Policy Distillation (RPD)~\cite{julg2025refined}
    uses a VLA teacher to guide the exploration of a compact,
    task-specific RL policy. Its training objective combines RL with
    a distillation loss that encourages agreement with the teacher's
    actions, integrating teacher guidance with reward-driven policy
    optimization.
    In \method, the teacher directly takes over when the student
    fails to make progress and executes a recovery trajectory.
    The student then learns from these recovery behaviors through
    imitation and distillation, without requiring RL
    optimization or task-specific reward design.

 \item RLDG~\cite{xu2024rldg} trains specialist policies with
    online RL and uses their rollouts as high-quality data to
    fine-tune a generalist robot policy.
    With the rapid development of EFMs,
    \method instead builds on the existing closed-loop control and
    recovery capabilities of EFM teachers, avoiding the need to train
    task-specific RL specialists.
    Moreover, our teacher-in-the-loop data collection in \ad
    provides recovery supervision tailored to states encountered
    by the student during execution.
    
\item PLD~\cite{xiao2026self} also collects recovery trajectories
    through takeover by task-specific residual policies trained with off-policy RL.
    However, its intervention timing is decided randomly, which does not explicitly account for the
    student policy's current execution progress.
    As shown in our analysis in \Cref{tab:intervention_timing}, premature takeover can produce
    trajectories that closely resemble offline demonstrations,
    providing limited additional recovery supervision.
    Conversely, excessively delayed intervention can allow errors to
    accumulate and lead to states from which recovery is difficult or
    impossible, reducing the efficiency of recovery-data collection.
    In \ad, the execution manager automatically triggers intervention
    based on task progress estimated by a general reward model.
    This allows teacher takeover to respond at critical steps,
    supporting targeted recovery-data collection and data-efficient
    policy improvement.
\end{itemize}

\section{Additional Implementation Details}
\label{supp:imple}

\subsection{AutoDAgger}
\label{supp:auto}

\textbf{Task Monitor.}
The robot client, policy server, and task monitor operate independently at
different frequencies. When the robot client
requests a new action chunk from the policy server, the corresponding front-camera
observation $\vo_n$ is also passed to the task monitor, where $n$ indexes policy-inference requests.
The task monitor appends this observation to a visual-history buffer
$\mathcal{B}_n$, which stores all frames received since the active
policy began execution.

As new observations arrive, the task monitor submits evaluation requests
to the reward model, subject to a minimum interval of
$\tau_{\mathrm{mon}}$ between consecutive submissions.
These requests are processed asynchronously, allowing policy inference and
robot execution to proceed without waiting for monitoring results.
Following the notation in \Cref{sec:ad}, $t$ indexes monitor updates within
the current policy execution segment, and $\tau_t$ denotes the time at
which the $t$-th monitor output is received. We use $n_t$ to denote the
index of the policy-inference request whose observation triggered the
corresponding monitor evaluation.

To limit the inference cost of the reward model, the task monitor samples at most
$N_{\mathrm{frame}}$ frames from the visual-history buffer. The sampled visual input for update $t$ is
\begin{equation}
    {\vo}_{t,1:K_t}
    = \operatorname{Sample}(\mathcal{B}_{n_t}, K_t),
    \quad
    K_t = \min\!\left\{|\mathcal{B}_{n_t}|, N_{\mathrm{frame}}\right\}.
    \label{eq:online_frame_sampling}
\end{equation}
If the buffer contains at most $N_{\mathrm{frame}}$ frames,
$\operatorname{Sample}(\cdot)$ returns all frames in chronological order.
Otherwise, it selects $N_{\mathrm{frame}}$ frames at evenly spaced indices
across the full history, including the most recent frame.

In our experiments, we use the official Robometer checkpoint\footnote{Hugging Face:
\texttt{robometer/Robometer-4B}} as the reward model without fine-tuning on downstream tasks.

\begin{table}[t]
    \centering
    \caption{Temporal thresholds used during \ad data collection.
    $T_{\mathrm{fail}}$ specifies the failure-detection timeout, while
    $T_{\mathrm{succ}}$ specifies the minimum time span of consecutive
    above-threshold monitor outputs required to confirm task success.}
    \small \label{tab:runtime_thresholds}
    \setlength{\tabcolsep}{7pt}
    \begin{tabular}{l|cc|cc}
        \toprule \multirow{2}{*}{Task} & \multicolumn{2}{c|}{$T_{\mathrm{fail}}$ (s)} & \multicolumn{2}{c}{$T_{\mathrm{succ}}$ (s)} \\
                                       & Student                                      & Teacher                                    & Student & Teacher \\
        \midrule Fold Dishcloth        & 6                                            & 14                                         & 5       & 5       \\
        Pick Pen                       & 3                                            & 8                                          & 3       & 3       \\
        Pick Beverage                  & 2                                            & 6                                          & 4       & 4       \\
        LIBERO & 5 & 10 & 4 & 4 \\
        \bottomrule
    \end{tabular}
\end{table}

\noindent
\textbf{Execution Manager.}
At the beginning of each episode, \ad clears the visual-history buffer and
prediction records, removes pending monitoring requests, and resets the
progress reference and timers.
For each policy execution segment, let $\tau_0$ denote its start time.
We initialize $\tau^{\mathrm{valid}}_0=\tau_0$.
Upon receiving the first monitor output, we set
$\bar{p}_1=p_1$, $g_1=1$, and
$\tau^{\mathrm{valid}}_1=\tau_1$.
Failure detection is enabled after this first update.
For $t\geq 2$, the manager updates the progress reference and the timestamp
of the most recent valid execution according to
\Cref{eq:progress_update,eq:valid_time_update}.

Between monitor updates, the manager uses the current wall-clock time to 
continuously track the elapsed time since $\tau^{\mathrm{valid}}_t$.
If the latest success probability satisfies $q_t\leq\theta_s$,
failure is detected as soon as this elapsed time reaches
$T_{\mathrm{fail}}$.
At monitor-update timestamps, this check reduces to
\Cref{eq:intervention_decision}.

The execution manager also determines whether the task has been completed.
When a monitor output first satisfies $q_t>\theta_s$, the manager starts
a success timer. The timer continues running as long as subsequent
predictions remain above the threshold. Any prediction at or below the
threshold resets the timer, which restarts at the next above-threshold
prediction.

Let $\tau_t^{\mathrm{succ}}$ denote the start time of the current success
timer. Success is checked only when a new monitor output arrives and is
confirmed when the latest prediction remains above the threshold and
at least $T_{\mathrm{succ}}$ has elapsed:
\begin{equation}
    q_t>\theta_s
    \;\land\;
    \tau_t-\tau_t^{\mathrm{succ}}\geq T_{\mathrm{succ}}.
    \label{eq:success_confirmation}
\end{equation}
Requiring consistent success predictions over this interval helps prevent
transient prediction spikes from being mistaken for task completion.

The execution manager polls the monitor associated with the active policy.
During student execution, the teacher monitor remains idle.
Upon detecting student failure, the manager immediately stops the robot,
clears the action buffer on robot controller, and transfers
control to the teacher.
The teacher monitor starts with a fresh visual history, and its update
index, progress reference, and timers are initialized as described above,
with $\tau_0$ set to the teacher takeover time.
The robot client resumes execution once the first teacher action chunk
becomes available.

During teacher recovery, the manager applies the same failure-detection
and success-confirmation rules using the teacher's monitoring state and
temporal thresholds.
If the teacher's success is confirmed, \ad retains the student rollout and
the subsequent teacher recovery segment as a single training trajectory.
If the teacher fails or the student completes the task without
intervention, the episode is discarded, as \ad specifically targets
the collection of successful recovery trajectories.

\noindent
\textbf{Hyperparameter Setup.}
Across all experiments, we fix $N_{\mathrm{frame}}=8$,
$\tau_{\mathrm{mon}}=1\,\mathrm{s}$, $\epsilon_p=0.05$, and
$\theta_s=0.5$.
The maximum number of sampled frames, $N_{\mathrm{frame}}$, balances
temporal context against monitoring cost, while $\tau_{\mathrm{mon}}$
controls the minimum interval between monitoring requests.
The progress margin $\epsilon_p$ prevents minor fluctuations in predicted
progress from being treated as meaningful advancement, and $\theta_s$
sets the threshold for predicted task success.

We configure the failure timeout $T_{\mathrm{fail}}$ and the
success-confirmation duration $T_{\mathrm{succ}}$ for each task,
as summarized in \Cref{tab:runtime_thresholds}.
In LIBERO tasks, time is measured in environment steps
and converted to seconds using the simulator's 10 FPS setup.
Within each setting, the student and teacher use different
temporal thresholds to accommodate their execution characteristics.

We assign shorter failure timeouts to the student to enable timely
teacher intervention and reduce the risk of further deviation from
recoverable states.
Longer teacher timeouts allow more time to recover from student-induced
off-nominal states.
The thresholds also vary across tasks to account for differences in
execution dynamics and expected duration.
For example, folding a dishcloth involves contact-rich manipulation and
may include temporary pauses during valid execution, whereas picking
tasks typically exhibit more visually apparent progress over shorter
timescales.

\subsection{Real-world Experiments}
\label{supp:real}

\begin{figure*}[t]
    \centering
    \includegraphics[width=0.8\textwidth]{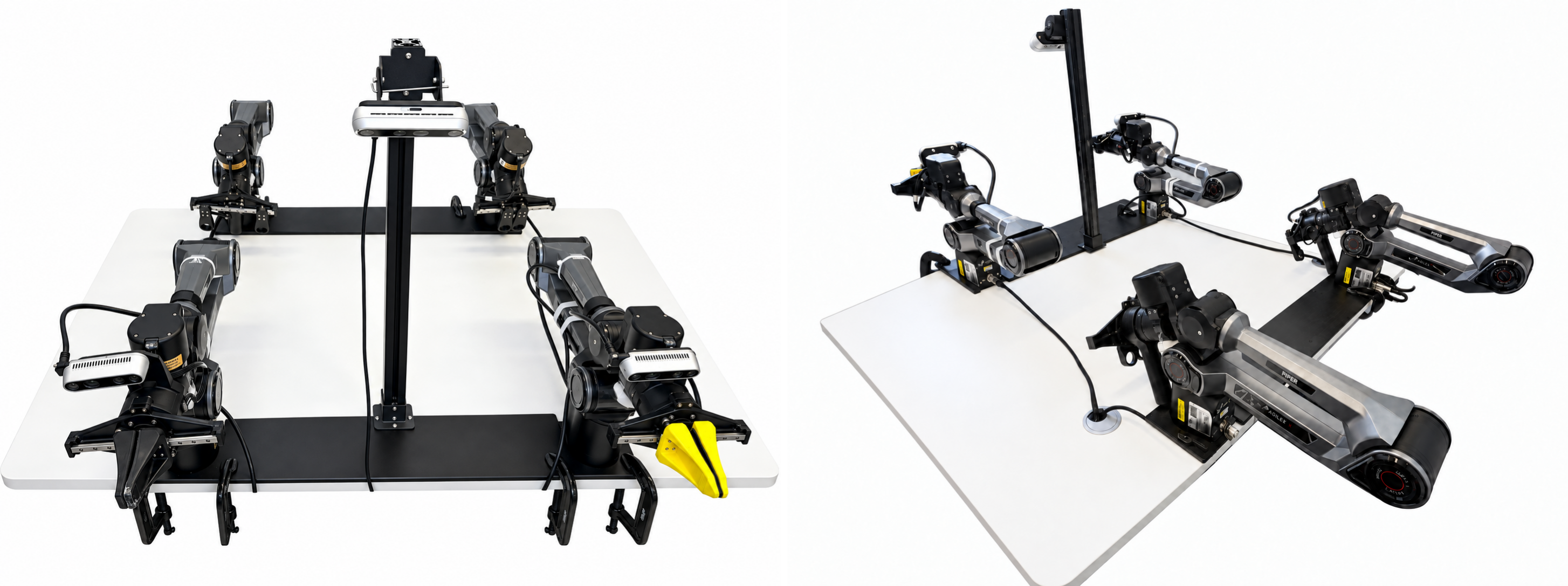}
    \caption{AgileX PiperX robotic arms.}
    \label{fig:piperx}
\end{figure*}

\noindent
\textbf{Hardware Setup.} We conduct our real-robot experiments on an AgileX PiperX
dual-arm robot platform, as illustrated in \Cref{fig:piperx}, following the leader-follower
teleoperation design of ALOHA-style systems~\citep{act, mobilealoha}. The platform
consists of four 7-DoF (6 joints + 1 gripper) PiperX arms: two leader arms
operated by a human demonstrator and two corresponding follower arms that
reproduce the leader commands during data collection and execute policy-generated
actions during autonomous inference. The visual sensing system comprises one front-facing
Intel RealSense D455 camera and two Intel RealSense D435 wrist cameras, with one
wrist camera mounted on each follower arm.

\begin{figure*}
    \centering
    \includegraphics[width=\linewidth]{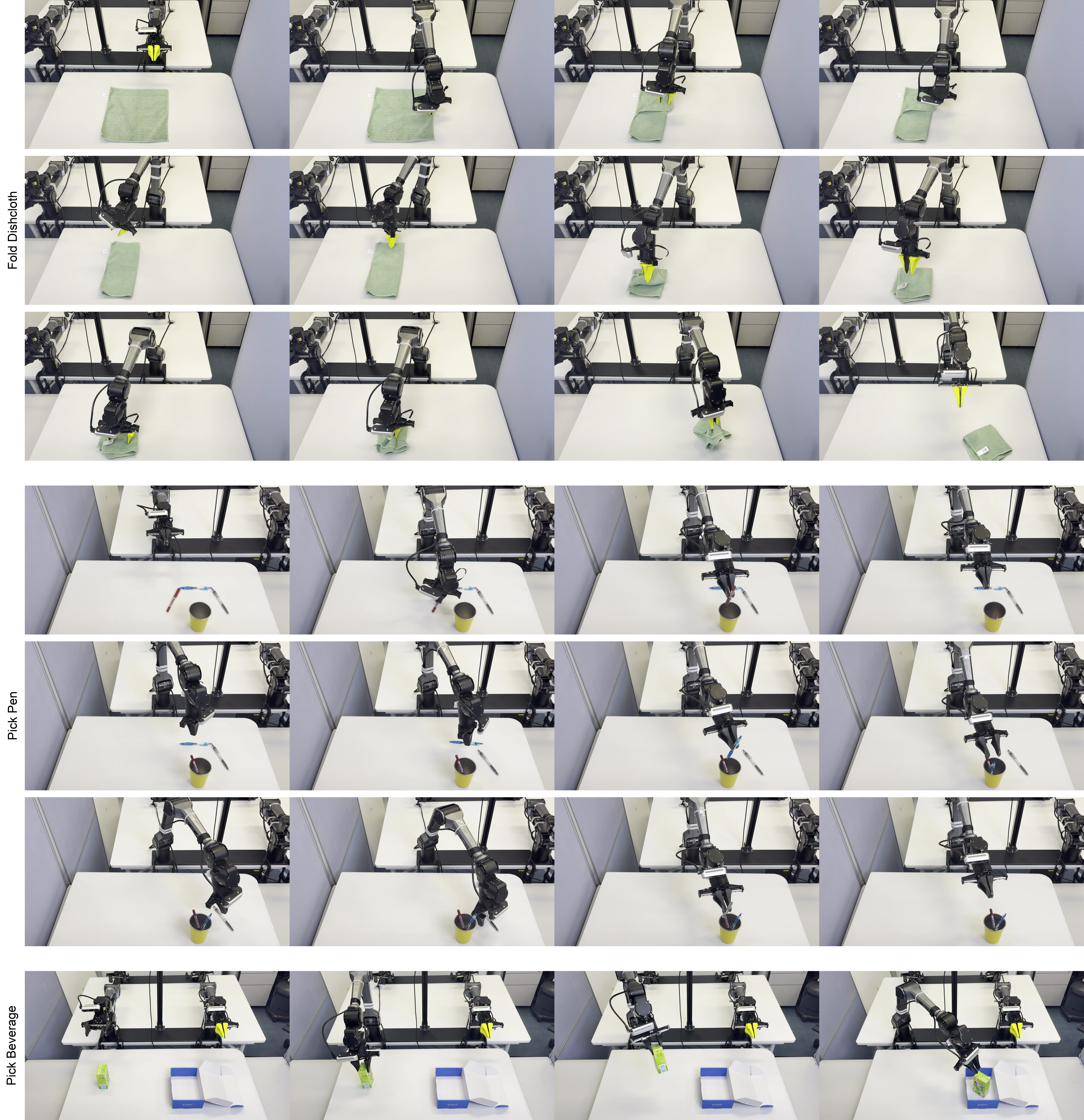}
    \caption{Visualization of three real-world tasks.}
    \label{fig:app_task}
\end{figure*}

\noindent
\textbf{Tasks.} We evaluate our method on three real-robot manipulation tasks: ``Fold
Dishcloth'', ``Pick Pen'', and ``Pick Beverage''. These tasks cover deformable-object
manipulation, precision grasping, and object relocation. Illustrations of the three
tasks are provided in \Cref{fig:app_task}.

The language instructions for the tasks are:
\begin{itemize}[noitemsep,topsep=0pt,leftmargin=*]
    \item \textbf{Fold Dishcloth:} ``Fold the dishcloth in half twice, then place
        it in the position slightly to the front and left.''

    \item \textbf{Pick Pen:} ``Pick up the scattered pens one by one and place
        them in the yellow cup.''

    \item \textbf{Pick Beverage:} ``Pick up the beverage and put it in the paper
        box.''
\end{itemize}

\noindent\textbf{Evaluation Protocol.}
Each method is evaluated using a fixed number of real-robot rollouts per task: 12
trials for Fold Dishcloth, 27 trials for Pick Pen, and 30 trials for Pick Beverage.
We use predefined test configurations, including fixed object categories and controlled
variations in object position and orientation, and keep these configurations as
consistent as practically possible across all evaluated methods. A rollout is considered
successful only if the complete task is accomplished within the maximum episode length
without human intervention.

In addition to binary task success, we report a fine-grained success score to
capture both partial task completion and execution quality under a limited real-robot
evaluation budget. Each task is decomposed into semantically meaningful stages.
Each stage receives a score of $0$, $0.5$, or $1$, corresponding to failure,
partial completion or successful completion after repeated attempts, and successful
completion in a single attempt, including immediate recovery from minor
execution issues, respectively. 
Subsequent stages that are not attempted due to failure at an earlier
stage are assigned a score of 0.
The task-level score is computed by averaging
over all stages and evaluation rollouts, and then multiplied by 100 for better display.
The scoring criteria are as follows:

\begin{enumerate}[noitemsep, topsep=0pt, leftmargin=*]
    \item Fold Dishcloth
        \begin{itemize}
            \item Stage 1: Folding the dishcloth in half from left to right
                \begin{itemize}
                    \item 0: The robot fails to complete the fold after multiple
                        attempts.

                    \item 0.5: The robot completes the fold after several
                        attempts or demonstrates partial recovery from an
                        execution error during the process.

                    \item 1: The robot completes the fold in a single attempt, including
                        cases in which it immediately recovers from minor issues
                        such as a flipped corner.
                \end{itemize}

            \item Stage 2: Folding the dishcloth in half a second time, from bottom
                to top
                \begin{itemize}
                    \item 0: The robot fails to complete the second fold after
                        multiple attempts.

                    \item 0.5: The robot completes the second fold after several
                        attempts or demonstrates partial recovery from an
                        execution error during the process.

                    \item 1: The robot completes the second fold in a single
                        attempt, including cases in which it immediately recovers
                        from minor issues such as a flipped corner.
                \end{itemize}

            \item Stage 3: Picking up the folded dishcloth and placing it in the
                front-left target region
                \begin{itemize}
                    \item 0: The robot fails to pick up and place the folded
                        dishcloth after multiple attempts.

                    \item 0.5: The robot completes the pick-and-place operation
                        after several attempts or demonstrates partial recovery
                        from an execution error during the process.

                    \item 1: The robot completes the pick-and-place operation in
                        a single attempt.
                \end{itemize}
        \end{itemize}

    \item Pick Pen
        \begin{itemize}
            \item Stage 1: Grasping the pen
                \begin{itemize}
                    \item 0: The robot fails to grasp the pen after multiple attempts. 

                    \item 0.5: The robot grasps the pen after repeated attempts
                        or recovers from an unsuccessful grasp.

                    \item 1: The robot grasps the pen correctly on the first
                        attempt.
                \end{itemize}

            \item Stage 2: Placing the pen into the cup
                \begin{itemize}
                    \item 0: The robot fails to place the pen into the cup after multiple attempts, or does not attempt recovery after the pen falls out of the cup. 

                    \item 0.5: The pen is placed into the cup after a recovery
                        attempt or with substantial contact against the cup.

                    \item 1: The robot directly places the pen into the cup
                        without requiring recovery.
                \end{itemize}

            \item For trials containing multiple pens, the two-stage scoring
                procedure is applied independently to each pen.
        \end{itemize}

    \item Pick Beverage
        \begin{itemize}
            \item Stage 1: Grasping the beverage
                \begin{itemize}
                    \item 0: The robot fails to grasp the beverage after multiple attempts.

                    \item 0.5: The robot grasps it after repeated attempts or
                        recovers from an unsuccessful grasp.

                    \item 1: The robot grasps it correctly on the first attempt.
                \end{itemize}

            \item Stage 2: Placing the beverage into the box
                \begin{itemize}
                    \item 0: The robot fails to place the beverage into the box after multiple attempts, or does not attempt recovery after the beverage falls out of the box. 

                    \item 0.5: The beverage is placed into the box after a
                        recovery attempt or with an undesirable collision.

                    \item 1: The robot directly places the beverage into the box
                        without requiring recovery.
                \end{itemize}
        \end{itemize}
\end{enumerate}

\subsection{Simulation Benchmarks}

\noindent\textbf{LIBERO.}
The LIBERO benchmark~\citep{libero} has four task suites---Spatial, Object, Goal,
and 10 (Long)---each emphasizing different manipulation capabilities. 
Each suite contains 10 tasks, with each task evaluated over
50 rollouts. We report the average task success rate for each suite.
    
For $\pi_{0.5}$, VLA-Adapter-Pro,
and SmolVLA, we use the dataset released by OpenVLA\footnote{HuggingFace: \texttt{openvla/modified\_libero\_rlds}}~\citep{openvla}
and convert it to the LeRobot format~\citep{lerobot}. For X-VLA, we instead use
the HDF5-format dataset released by the X-VLA authors\footnote{HuggingFace: \texttt{2toINF/Libero-XVLA-format}},
which represents actions in the EE6D action space.

\noindent\textbf{LIBERO-Plus.}
LIBERO-Plus~\citep{libero_plus} extends LIBERO with seven dimensions of visual
and configuration perturbations. In our pilot study (\Cref{fig:teaser}), we
evaluate the ``Background'' setting, which perturbs scene and surface appearance,
and the ``Robot'' setting, which varies the manipulator's initial pose. These
settings contain 1,076 and 1,550 task instances, respectively. The evaluations
are performed using checkpoints trained exclusively on the original LIBERO data,
thereby measuring out-of-distribution generalization under environmental and embodiment
perturbations.

\subsection{Data Collection}
Our offline dataset contains 200 human demonstration
episodes for Fold Dishcloth and 100 episodes each for Pick Pen and
Pick Beverage. For LIBERO experiments, we use the standard training set
described above.

For interactive data collection, we first train the student policy on
$\mathcal{D}_{\mathrm{Demo}}$ using behavior cloning to obtain the BC
baseline. We then use this same fixed baseline policy to collect data
with both AutoDAgger and HG-DAgger, ensuring a fair comparison.
After collection, we reinitialize the student policy and train it on
the corresponding aggregated dataset.

\subsection{Training Setup}
We train $\pi_{0.5}$\footnote{GitHub: \texttt{Physical-Intelligence/openpi}}, X-VLA\footnote{GitHub:
\texttt{2toinf/X-VLA}}, VLA-Adapter-Pro\footnote{GitHub: \texttt{OpenHelix-Team/VLA-Adapter}},
and SmolVLA\footnote{GitHub: \texttt{huggingface/lerobot}} based on their
official codebases, with modifications to support our training pipeline and data
formats. 
We follow the default or recommended training configurations for each model, with
the key hyperparameters summarized in \Cref{tab:hyperparam}.

For real-robot experiments, demonstrations are recorded in absolute joint
space at 30FPS. When required, absolute joint actions are
converted to relative joint actions by subtracting the corresponding
proprioceptive states.
For LIBERO experiments, the policies in \Cref{tab:cross_smolvla} are trained
jointly on all four suites, whereas those in
\Cref{tab:intervention_timing} are trained only on LIBERO-10 to focus the
analysis on the most challenging tasks.

During training, we randomly sample from the pooled training datasets
without additional dataset balancing. The loss applied to each sample
is determined by its data source. The supervision used for each training
variant is summarized in \Cref{tab:config}.

\begin{table}[t]
    \centering
    \caption{Details of different training configurations.}
    \label{tab:config} \small
    \setlength{\tabcolsep}{12pt}
    \begin{tabular}{l|ccc}
        \toprule Method      & $\mathcal{D}_{\mathrm{Demo}}$ & $\mathcal{D}_{\mathrm{Roll}}$ & $\mathcal{D}_{\mathrm{Rec}}$ \\
        \midrule BC Baseline & BC                            & --                            & --                           \\
        Offline Distill      & BC+KD                         & --                            & --                           \\
        OPD                  & BC+KD                         & KD                            & --                           \\
        Recovery BC          & BC                            & --                            & BC                           \\
        Recovery Distill     & BC+KD                         & --                            & BC+KD                        \\
        \method              & BC+KD                         & KD                            & BC+KD                        \\
        \bottomrule
    \end{tabular}
\end{table}

During distillation, when both the teacher and student use the
$\pi_{0.5}$ architecture, we compute the distillation loss
$\mathcal{L}_{\mathrm{KD}}(\vs,\vl)$ by directly matching their
flow-matching velocity predictions using shared flow matching timesteps and noise samples.
When the student uses a different architecture, such as X-VLA or SmolVLA,
we use the teacher's fully denoised action predictions as supervision
targets after post-processing and denormalization into the native action
space. The student is then trained on these targets using the same
imitation learning objective as for ground-truth actions.
We set the distillation loss weight $\lambda_{\mathrm{KD}}=1$
in all experiments.

All models are trained on 8 NVIDIA A800 GPUs, each with 80GB VRAM. Training takes
approximately 1--2 days for $\pi_{0.5}$ and less than one day for the other models.

\begin{table*}[t]
    \centering
    \caption{Hyperparameter settings for training $\pi_{0.5}$, X-VLA, VLA-Adapter-Pro, and SmolVLA policies.}
    \label{tab:hyperparam}
    \small
    \setlength{\tabcolsep}{6.5pt}
    \renewcommand{\arraystretch}{1.2}
        \begin{tabular}{c|ccccccc}
            \toprule
            Model & \multicolumn{2}{c}{$\pi_{0.5}$} & \multicolumn{2}{c}{X-VLA} & \multicolumn{2}{c}{VLA-Adapter-Pro} & SmolVLA \\
            Task & AgileX & LIBERO & AgileX & LIBERO & AgileX & LIBERO & LIBERO \\
            \midrule
            Prediction Horizon $H$ & 50 & 10 & 30 & 30 & 30 & 8 & 10 \\
            Action Space & Relative Joint & Delta EEF & Abs Joint & Abs EE6D & Abs Joint & Delta EEF & Delta EEF \\
            Global Batch Size & 128 & 256 & 128 & 128 & 128 & 64 & 64 \\
            Training Steps & 50k & 30k & 50k & 60k & 50k & 100k & 50k \\
            Optimizer & AdamW & AdamW & AdamW & AdamW & AdamW & AdamW & AdamW \\
            Weight Decay & 0 & 0 & 0 & 0 & 0.01 & 0.01 & 1e-10 \\
            Betas & (0.9, 0.95) & (0.9, 0.95) & (0.9, 0.95) & (0.9, 0.95) & (0.9, 0.999) & (0.9, 0.999) & (0.9, 0.95) \\
            Base LR & 2.5e-5 & 5e-5 & 1e-4 & 1e-4 & 2e-4 & 2e-4 & 1e-4 \\
            LR Scheduler & Cosine Decay & Cosine Decay & Constant & Constant & Constant & Constant & Cosine Decay \\
            Warmup Steps & 1k & 10k & 2k & 2k & 0 & 0 & 1k \\
            Grad Norm Clip & 1.0 & 1.0 & 1.0 & 1.0 & N/A & N/A & 10.0 \\
            EMA Decay & 0.99 & 0.999 & N/A & N/A & N/A & N/A & N/A \\
            \bottomrule
        \end{tabular}
\end{table*}

\subsection{Robot Deployment Setup}
We deploy the policy and task monitor models on a workstation equipped with an
NVIDIA RTX 4090 GPU. The inference latency is
$80.0\pm1.6$ ms for $\pi_{0.5}$ and
$266.9\pm0.5$ ms for Robometer.
The robot client runs on ROS and executes actions at a fixed control
frequency of $30$ Hz.
We combine asynchronous inference with temporal ensembling to maintain
continuous control and smooth action trajectories.

\subsection{Execution Manager Experiments}

We evaluate failure detection in \Cref{tab:intervention_acc} on all
10 LIBERO-10 tasks, using 20 initial states per task and an intermediate
SmolVLA checkpoint as the student policy.
We first collect student-only trajectories and then evaluate failure
detection against their ground-truth outcomes, as determined by the
environment's criterion.

To assess intervention timing in \Cref{tab:intervention_timing},
we replay each recorded student trajectory up to the takeover step
$\max(0,s_{\mathrm{AD}}+\Delta)$ and then transfer control to the teacher.
Here, $s_{\mathrm{AD}}$ denotes the intervention step selected by execution manager,
and $\Delta\in\{-50,-20,0,20,50\}$ shifts the takeover earlier or later.
We also compare against fixed takeover steps of 50, 100, and 200,
as well as a random step sampled uniformly from $[0,520]$
for each episode, where 520 is the maximum episode length.

The relative-offset comparisons are restricted to the 136 episodes
in which \ad triggers intervention. The fixed-step and random
baselines are evaluated on the full set of 200 student trajectories.
Episodes ending before the scheduled takeover retain their
student outcomes.
For all strategies, recovery success rates are computed only over
episodes in which teacher takeover actually occurs.

For distilled student performance, we randomly sample 50 episodes for each
strategy and train the student policy from SmolVLA w/o PT using \method for
fair comparison.

\section{Additional Experimental Results}
\label{supp:exp}

\subsection{Validation of Execution Manager}

\Cref{tab:failure_10} provides the full confusion matrix for the
LIBERO-10 failure-detection results reported in
\Cref{tab:intervention_acc}.
We further evaluate on 20 real-robot rollouts
of the Fold Dishcloth task, consisting of 10 student rollouts and
10 teacher rollouts.
Ground-truth outcomes are manually annotated according to whether
the rollout successfully completes the task.
The student rollouts contain two successes and eight failures,
while the teacher rollouts contain eight successes and two failures.
The combined evaluation set is therefore balanced, with
10 successful and 10 failed trajectories.
Consistent with \Cref{tab:intervention_acc} and \Cref{tab:failure_10}, we treat failure as
the positive class.

As shown in \Cref{tab:failure_real}, the execution manager correctly
flags 9 of the 10 failed trajectories and achieves a precision of $81.8\%$, a recall of $90.0\%$,
and an F1 score of $85.7\%$.
The two unnecessary interventions (FP) indicate that the monitor adopts a conservative criterion for confirming
task success, and only one of 10 failed episodes is missed (FN).
These results provide evidence that the execution manager
can supply useful failure-detection signals for teacher takeover
and recovery monitoring in real-world execution.

\begin{table}[t]
    \centering
    \caption{\textbf{Failure detection by the execution manager on
    LIBERO-10.} The confusion matrix supplements the results in
    \Cref{tab:intervention_acc}. }
    \label{tab:failure_10}
    \setlength{\tabcolsep}{13pt}
    \small
    \begin{tabular}{l|cc|c}
        \toprule \multirow{2}{*}{Prediction} & \multicolumn{2}{c|}{Ground-truth Outcome} & \multirow{2}{*}{Total} \\
                                             & Failure                                      & Success                 &    \\
        \midrule Triggered                    & 124 (TP)                                    & 12 (FP)               & 136  \\
        Not Triggered                             & 1 (FN)                                  & 63 (TN)             & 64 \\
        \midrule Total                       & 125                                        & 75                    & 200 \\
        \bottomrule
    \end{tabular}
\end{table}

\begin{table}[t]
    \centering
    \caption{\textbf{Failure detection by the execution manager on
    Fold Dishcloth.} Predictions are evaluated against
    manually annotated ground truth, with failure treated
    as the positive class.}
    \label{tab:failure_real}
    \setlength{\tabcolsep}{13.5pt}
    \small
    \begin{tabular}{l|cc|c}
        \toprule \multirow{2}{*}{Prediction} & \multicolumn{2}{c|}{Ground-truth Outcome} & \multirow{2}{*}{Total} \\
                                             & Failure                                      & Success                 &    \\
        \midrule Triggered                    & 9 (TP)                                     & 2 (FP)               & 11  \\
        Not Triggered                             & 1 (FN)                                 & 8 (TN)              & 9 \\
        \midrule Total                       & 10                                        & 10                    & 20 \\
        \bottomrule
    \end{tabular}
\end{table}

\subsection{Experimental Result Statistics}

We supplement the performance statistics of experiments in the main text in \Cref{tab:stats1},
\Cref{tab:stats2}, \Cref{tab:stats3}, and \Cref{tab:stats4}.

\subsection{Visualization of \ad Collection}

\begin{figure*}
    \centering
    \includegraphics[width=0.95\linewidth]{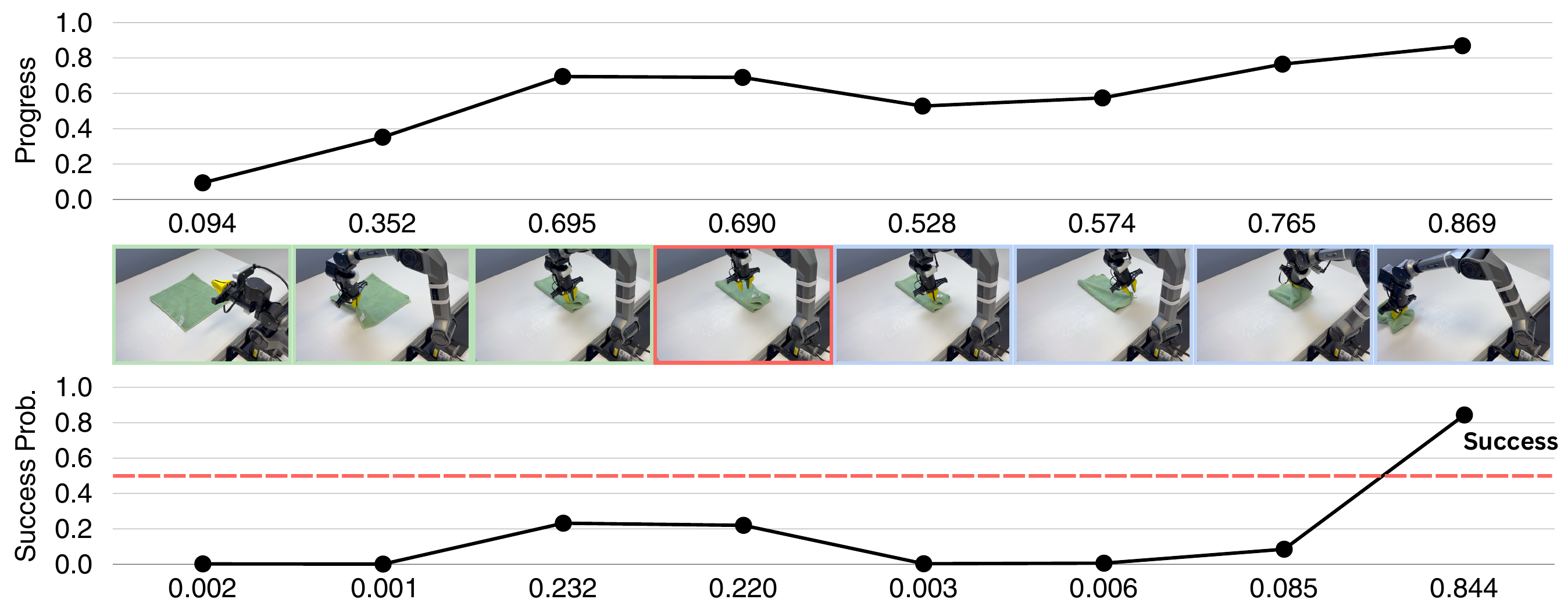}
    \vspace{1mm}

    \includegraphics[width=0.95\linewidth]{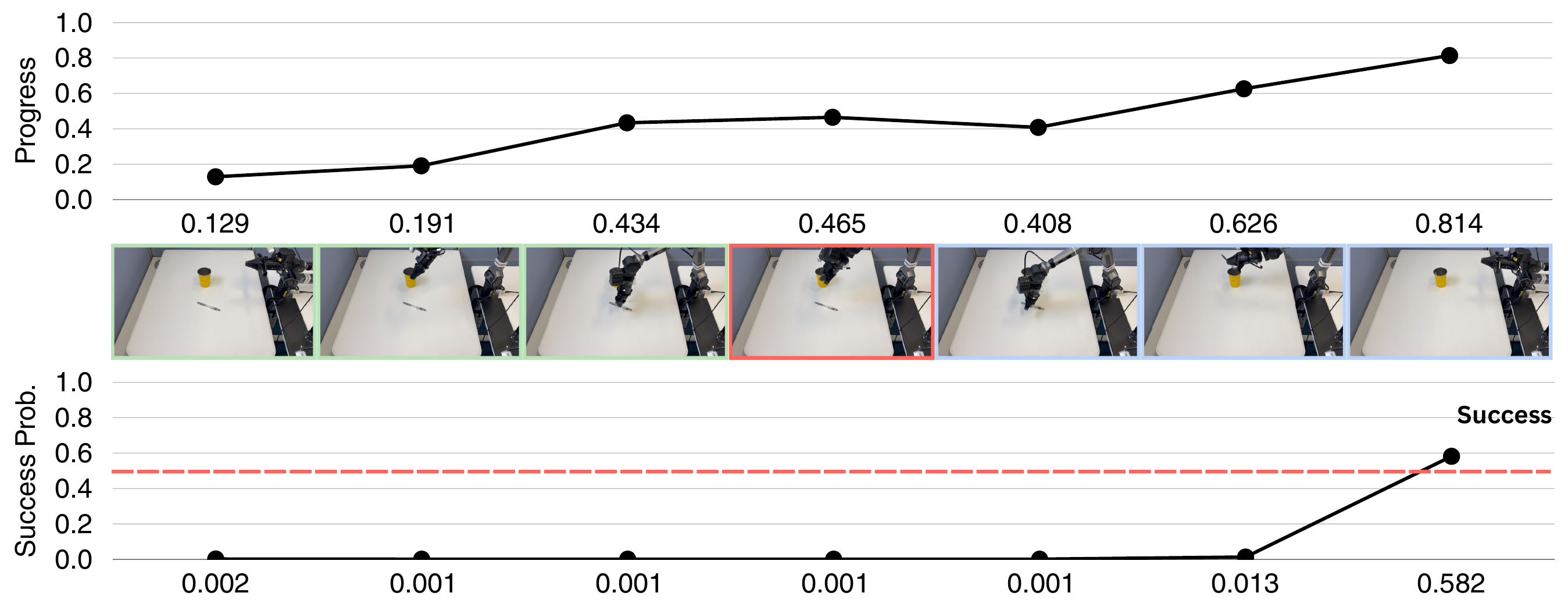}
    \vspace{1mm}

    \includegraphics[width=0.95\linewidth]{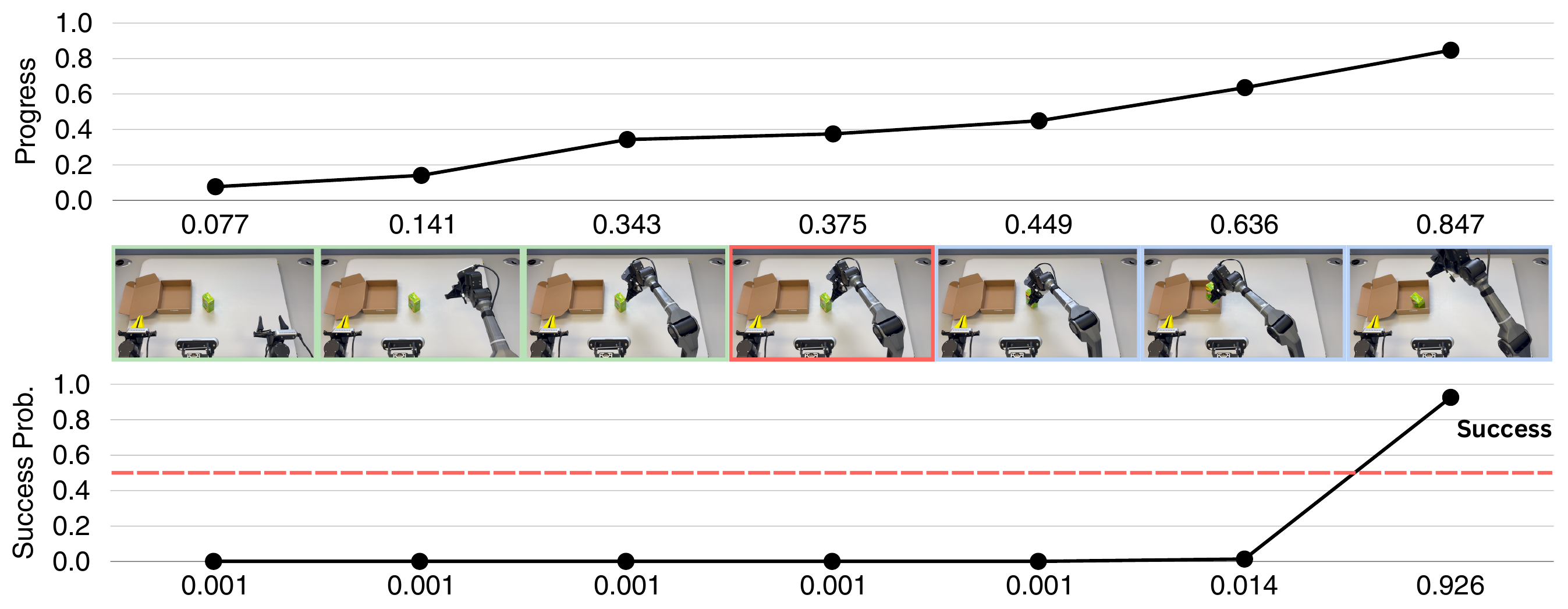}
    \caption{\textbf{Representative AutoDAgger collection trajectories.} Examples
    for Fold Dishcloth, Pick Pen, and Pick Beverage are shown from top to bottom.
    Green, red, and blue frame borders denote student execution, detected failure,
    and teacher recovery, respectively. The red dashed line marks the success
    threshold $\theta_{s}=0.5$. For clarity, the figure shows only selected frames rather
than the full sequence of consecutive frames.}
    \label{fig:app_ad_task}
\end{figure*}

We visualize representative AutoDAgger collection trajectories for Fold Dishcloth,
Pick Pen, and Pick Beverage tasks in \Cref{fig:app_ad_task}. Before intervention,
the predicted task progress initially increases during student execution but subsequently
stagnates, while the success probability remains well below $\theta_{s}=0.5$. The
execution manager consequently detects the combination of sustained progress stagnation
and low success probability, triggering teacher takeover at the red-bordered
frames. Following takeover, the predicted progress may decrease temporarily as
the teacher corrects the student-induced state. It subsequently resumes increasing
as recovery proceeds, reaching $0.869$, $0.814$, and $0.847$ for the three tasks.
Meanwhile, the success probability remains low during most of the recovery
process and rises above $\theta_{s}$ only near task completion, reaching $0.844$,
$0.582$, and $0.926$, respectively.

\section{Limitations and Future Work}
\label{supp:limitation}

Despite the improved student performance demonstrated by \method, several
limitations remain. First, TILDE is evaluated with a single round of AutoDAgger
collection followed by student training, rather than iterative data aggregation.
Extending the experiments to alternate between policy updates and subsequent
rounds of AutoDAgger could progressively expose the student to increasingly
relevant failure and recovery states. Second, AutoDAgger depends on the teacher's
ability to recover from failure states. States beyond the teacher's recovery capabilities
can reduce the yield of effective data and collection efficiency. Following
common assumption in interactive imitation learning, settings with irreversible failures,
such as dynamic tasks, fall outside the scope of this work. Finally, our experimental
evaluation focuses on a limited set of real-world tasks, simulation benchmarks, and
teacher--student architectures. Future work will scale \method to broader tasks,
embodiments, and heterogeneous policy families, and investigate how closely
student policies can approach pretrained EFMs with continued autonomous data
aggregation and distillation.

\begin{table*}
    [h]
    \centering
    \caption{Performance statistics for the real-world tasks, corresponding to \Cref{fig:main_results}
    and \Cref{fig:ablation}. The 96\% confidence intervals are computed using bootstrap
    resampling.}
    \label{tab:stats1} \small
    \setlength{\tabcolsep}{18pt}
    \begin{tabular}{c|cccccc}
        \toprule \multirow{2}{*}{Method} & \multicolumn{2}{c}{Fold Dishcloth} & \multicolumn{2}{c}{Pick Pen} & \multicolumn{2}{c}{Pick Beverage} \\
                                         & Mean                               & 96\% CI                      & Mean                             & 96\% CI      & Mean & 96\% CI      \\
        \midrule Teacher                 & 87.5                               & (69.4, 98.6)                 & 87.0                             & (78.7, 94.4) & 79.2 & (65.0, 91.7) \\
        BC Baseline                      & 20.8                               & (2.8, 41.7)                  & 37.0                             & (22.2, 51.9) & 16.7 & (8.3, 25.8)  \\
        Offline Distill                  & 27.8                               & (6.9, 51.4)                  & 45.4                             & (31.5, 60.2) & 26.7 & (14.2, 40.0) \\
        OPD                              & 43.1                               & (16.7, 69.4)                 & 58.3                             & (45.4, 70.4) & 40.0 & (25.0, 55.0) \\
        Recovery BC                      & 48.6                               & (22.2, 75.0)                 & 63.0                             & (48.1, 76.9) & 50.0 & (32.5, 67.5) \\
        Recovery Distill                 & 54.2                               & (27.8, 80.6)                 & 72.2                             & (56.5, 86.1) & 68.3 & (54.2, 80.8) \\
        TILDE                            & 63.9                               & (38.9, 87.5)                 & 80.6                             & (72.2, 88.0) & 69.2 & (53.3, 83.3) \\
        \bottomrule
    \end{tabular}
\end{table*}

\begin{table*}
    [h]
    \centering
    \caption{Performance statistics for the real-world tasks corresponding to
    \Cref{tab:data_source}.}
    \label{tab:stats2}
    \setlength{\tabcolsep}{14pt}
    \small
    \begin{tabular}{lc|cccccc}
        \toprule \multirow{2}{*}{Data Source} & \multirow{2}{*}{Distill} & \multicolumn{2}{c}{Fold Dishcloth} & \multicolumn{2}{c}{Pick Pen} & \multicolumn{2}{c}{Pick Beverage} \\
                                              &                          & Mean                               & 96\% CI                      & Mean                             & 96\% CI      & Mean & 96\% CI      \\
        \midrule Human Demo                   & \xmark                   & 20.8                               & (2.8, 41.7)                  & 37.0                             & (22.2, 51.9) & 16.7 & (8.3, 25.8)  \\
        \quad +Extra Demo                     & \xmark                   & 27.8                               & (6.9, 51.4)                  & 45.4                             & (27.8, 63.0) & 44.2 & (28.3, 60.0) \\
        \quad +\ad                            & \xmark                   & 48.6                               & (22.2, 75.0)                 & 63.0                             & (48.1, 76.9) & 50.0 & (32.5, 67.5) \\
        \midrule Human Demo                   & \cmark                   & 27.8                               & (6.9, 51.4)                  & 45.4                             & (31.5, 60.2) & 26.7 & (14.2, 40.0) \\
        \quad +Extra Demo                     & \cmark                   & 29.2                               & (9.7, 51.4)                  & 54.6                             & (40.7, 68.5) & 48.3 & (33.3, 63.3) \\
        \quad +\ad                            & \cmark                   & 63.9                               & (38.9, 87.5)                 & 80.6                             & (72.2, 88.0) & 69.2 & (53.3, 83.3) \\
        \bottomrule
    \end{tabular}
\end{table*}

\begin{table*}
    [h]
    \centering
    \caption{Performance statistics for the real-world tasks corresponding to
    \Cref{tab:human_vs_auto}.}
    \label{tab:stats3} \small
    \setlength{\tabcolsep}{14.9pt}
    \begin{tabular}{lc|cccccc}
        \toprule \multirow{2}{*}{Method}     & \multirow{2}{*}{Distill} & \multicolumn{2}{c}{Fold Dishcloth} & \multicolumn{2}{c}{Pick Pen} & \multicolumn{2}{c}{Pick Beverage} \\
                                             &                          & Mean                               & 96\% CI                      & Mean                             & 96\% CI      & Mean & 96\% CI      \\
        \midrule \multirow{2}{*}{HG-DAgger}  & \xmark                   & 34.7                               & (11.1, 59.7)                 & 75.9                             & (60.2, 89.8) & 60.0 & (46.7, 73.3) \\
                                             & \cmark                   & 65.3                               & (33.3, 90.3)                 & 80.6                             & (68.5, 90.7) & 63.3 & (48.3, 77.5) \\
        \midrule \multirow{2}{*}{AutoDAgger} & \xmark                   & 48.6                               & (22.2, 75.0)                 & 63.0                             & (48.1, 76.9) & 50.0 & (32.5, 67.5) \\
                                             & \cmark                   & 63.9                               & (38.9, 87.5)                 & 80.6                             & (72.2, 88.0) & 69.2 & (53.3, 83.3) \\
        \bottomrule
    \end{tabular}
\end{table*}

\begin{table*}
    \centering
    \caption{Recovery data collection statistics for the real-world tasks
    corresponding to \Cref{tab:human_vs_auto}. $T_\mathrm{Rec}$ is reported as mean
    $\pm$ standard deviation.}
    \label{tab:stats4} \small
    \setlength{\tabcolsep}{11.9pt}
    \begin{tabular}{l|ccc|ccc}
        \toprule \multirow{2}{*}{Task} & \multicolumn{3}{c|}{HG-DAgger} & \multicolumn{3}{c}{AutoDAgger} \\
                                       & $T_\mathrm{Rec}$      & Rec. SR (\%)                  & 96\% CI       & $T_\mathrm{Rec}$ & Rec. SR (\%) & 96\% CI       \\
        \midrule Fold Dishcloth        & $54.4 \pm 10.9$s               & 76.7                          & (60.0, 90.0)  & $21.5 \pm 5.8$s & 73.3         & (56.7, 90.0)  \\
        Pick Pen                       & $29.4 \pm 2.4$s                & 76.7                          & (60.0, 90.0)  & $13.2 \pm 1.4$s & 66.7         & (50.0, 83.3)  \\
        Pick Beverage                  & $32.2 \pm 2.8$s                & 90.0                          & (76.7, 100.0) & $12.4 \pm 1.8$s & 90.0         & (76.7, 100.0) \\
        \bottomrule
    \end{tabular}
\end{table*}

\end{document}